%% file: main.tex
\documentclass[preprint,review,3p,12pt,times,sort&compress]{elsarticle}

\usepackage{amsmath,amssymb}

\usepackage{algorithm}
\usepackage{algorithmic}
\usepackage{multirow}
\usepackage{newfloat}
\usepackage{listings}
\usepackage{booktabs}
\usepackage{bbding}
\usepackage{lineno,hyperref}
\usepackage{subfigure}
\usepackage{tabularx}
\usepackage{caption}
\usepackage{color}
\usepackage{hyperref}
\modulolinenumbers[5]

\renewcommand{\textcolor}[2]{#2}

\journal{Pattern Recognition}

\begin{document}

\begin{frontmatter}



\title{FILP-3D: Enhancing 3D Few-shot Class-incremental Learning with Pre-trained Vision-Language Models}

\author[label1]{Wan Xu}
\author[label1]{Tianyu Huang}
\author[label1]{Tianyuan Qu}
\author[label1]{Guanglei Yang\corref{mycorrespondingauthor}}
\author[label2]{Yiwen Guo}
\author[lable1]{Wangmeng Zuo}
\cortext[mycorrespondingauthor]{Corresponding author}
\affiliation[label1]{addressline={Harbin Institute of Technology},
            city={Harbin},
            country={China}}

\affiliation[label2]{organization={Independent Researcher}}
\ead{yangguanglei@hit.edu.cn}



\input{sections/0_abstract}
\begin{keyword}
3D classification, FSCIL, 3D from multi-view, V-L pre-trained model
\end{keyword}








\end{frontmatter}

\input{sections/1_introduction}
\input{sections/2_related_work}
\input{sections/2.5_symbol_table}
\input{sections/3_method}
\input{sections/4_experiment}
\input{sections/5_conclusion}
\bibliographystyle{elsarticle-num-names}
\bibliography{reference}

\end{document}

%% file: sections/0_abstract.tex
\begin{abstract}
Few-shot class-incremental learning (FSCIL) aims to mitigate the catastrophic forgetting issue when a model is incrementally trained on limited data. \textcolor{red}{However, many of these works lack effective exploration of prior knowledge, rendering them unable to effectively address the domain gap issue in the context of 3D FSCIL, thereby leading to catastrophic forgetting. The Contrastive Vision-Language Pre-Training (CLIP) model serves as a highly suitable backbone for addressing the challenges of 3D FSCIL due to its abundant shape-related prior knowledge. Unfortunately, its direct application to 3D FSCIL still faces the incompatibility between 3D data representation and the 2D features, primarily manifested as feature space misalignment and significant noise. To address the above challenges, we introduce the FILP-3D framework with two novel components: the Redundant Feature Eliminator (RFE) for feature space misalignment and the Spatial Noise Compensator (SNC) for significant noise.} RFE aligns the feature spaces of input point clouds and their embeddings by performing a unique dimensionality reduction on the feature space of pre-trained models (PTMs), effectively eliminating redundant information without compromising semantic integrity. On the other hand, SNC is a graph-based 3D model designed to capture robust geometric information within point clouds, thereby augmenting the knowledge lost due to projection, particularly when processing real-world scanned data. Moreover, traditional accuracy metrics are proven to be biased due to the imbalance in existing 3D datasets. Therefore we propose 3D FSCIL benchmark FSCIL3D-XL and novel evaluation metrics that offer a more nuanced assessment of a 3D FSCIL model. Experimental results on both established and our proposed benchmarks demonstrate that our approach significantly outperforms existing state-of-the-art methods. Code is available at: \url{https://github.com/HIT-leaderone/FILP-3D}

\end{abstract}

%% file: sections/1_introduction.tex
\section{Introduction}
The advancement of deep learning has led to significant progress in class recognition within predefined label sets, as demonstrated by various groundbreaking models~\cite{ViT, PointNet, ULIP, zhang2023controlvideo}. 
However, in real-world settings, it's common to encounter new types of data in a continuous stream, introducing classes that were not part of the original training. 
This situation is especially challenging in areas where collecting data is expensive, such as identifying rare animals~\cite{FACT}. 
In response, the research community has introduced several Few-Shot Class-Incremental Learning (FSCIL) models~\cite{FSCIL, BiDist, FSCIL-3D}. 
These models aim to gradually learn from a small number of examples of new concepts while keeping the knowledge previously gained. 
Although much of the existing research ~\cite{FSCIL, BiDist} has focused on image data, these approaches often fall short for FSCIL tasks in point cloud data. 
For example, a robot that has been trained extensively on synthetic data needs to adjust to new classes based on real-world point clouds. 
In this scenario, the domain gap between synthetic training data and real-world data makes feature-prototype alignments cumbersome during the incremental steps, eventually magnifying forgetting and overfitting issues~\cite{FSCIL-3D}. 
Therefore, there is a critical need for 3D FSCIL models that not only tackle the issues of catastrophic forgetting and overfitting but also adapt to the domain gap encountered during incremental learning.
The pioneering work by ~\citet{FSCIL-3D} marks the first significant exploration into 3D FSCIL, offering a novel solution to this issue.
They propose extracting principal components from base task instances, termed microshapes, which represent distinct aspects of 3D shapes.
This method allows for the effective alignment of 3D point cloud features with the language prototypes of classes, facilitating more accurate class recognition. 
\textcolor{red}{However, the microshapes primarily draw from knowledge of base task instances and lack supplementation of sufficient shape-related prior knowledge. Therefore, the features exacted by their model lack sufficient generalization capability. Moreover, the point features extracted by PointNet~\cite{PointNet} are not in alignment with the textual prototypes. 
Thus, ~\cite{FSCIL-3D} finds it challenging to simultaneously retain features associated with base classes (old knowledge) and integrate features of novel classes (new knowledge). In contrast, several methods~\cite{zhu2023not,kang2024bibimbap} have shown that the extensive prior knowledge embedded in Vision-Language (V-L) Pre-trained Models (PTMs) can significantly bridge the domain gap in domain adaptation tasks. Furthermore, Radford et al.~\cite{CLIP1} demonstrate that the general and adaptable embeddings provided by V-L PTMs are effective in addressing the dual challenges of forgetting and overfitting commonly encountered in FSCIL tasks. Given above insights, V-L PTMs emerge as highly promising backbones for overcoming the specific challenges associated with 3D FSCIL.
Therefore, we attempt to introduce PTMs into the 3D FSCIL task, to address the lack of prior knowledge and the issue of naturally aligned encoders in previous 3D FSCIL works~\cite{FSCIL-3D}.}

\begin{figure}[t]
  \centering
  \includegraphics[width=0.8\textwidth]{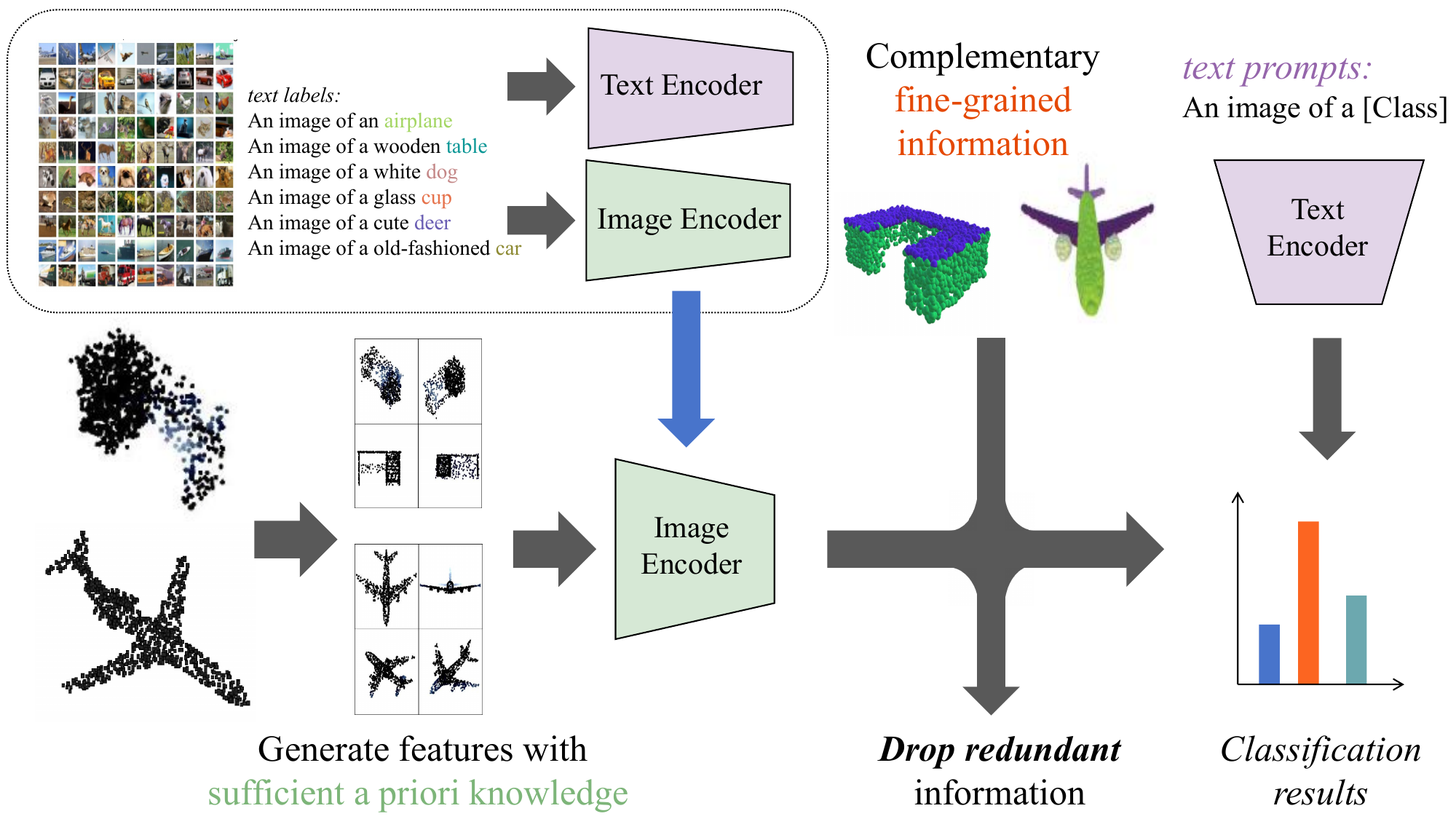}
  \caption{CLIP learns a large amount of prior knowledge from massive image-text pairs. Thus pre-aligned image and text features contain sufficient shape-related prior knowledge. Along with the elimination of redundant information (RFE) and the compensator of 3D fine-grained information (SNC), the performance in 3D FSCIL can be significantly improved.}
  \label{fig:teaser}
  \vspace{-2em}
\end{figure}

The incorporation of V-L PTMs into Few-Shot Class-Incremental Learning (FSCIL) tasks, through direct projection techniques, has marked a significant advancement over previous methodologies according to our experiment (please refer to Section.~6 for more details). 
\textcolor{red}{Despite these improvements, the incompatibility between 3D data representation and the 2D feature encoder impede further performance enhancements, primarily manifested as two aspects.
First, point cloud data, which consists of discrete points in 3D space, primarily represents geometric features of objects. It inherently misses richer visual nuances, such as color and texture. This deficiency is especially significant given CLIP's visual encoder, which inherently seeks and incorporates these visual details into its semantic representation~\cite{CLIP}. Consequently, this feature space misalignment could harm the precise classification of depth maps.
Second, point clouds derived from real-world scans often suffer from significant noise~\cite{co3d}, which distorts the extraction of depth-centric features. This problem is accentuated in multi-view projection techniques, where a significant fraction of the points might be occluded, thereby increasing the potential noise~\cite{zhao2021point}. Such disturbances can enlarge the domain gap and upset the balance in few-shot incremental learning, leading to overfitting~\cite{FSCIL-3D}. }

\textcolor{red}{To tackle the mismatch between 3D representation and V-L PTM's 2D encoder in direct projection techniques~\cite{FSCIL-3D, zhang2022pointclip}, we introduce \textbf{F}ew-shot class \textbf{I}ncremental \textbf{L}earning tasks with \textbf{P}re-training on \textbf{3D} (FILP-3D), a framework that employs CLIP as its backbone and incorporates two innovative components: Redundant Feature Eliminator (RFE) to tackle the feature space misalignment and Spatial Noise Compensator (SNC) to tackle the significant noise issue, as is shown in Figure~\ref{fig:teaser}.} RFE serves as a specialized dimensionality reduction technique, designed to remove redundant features while preserving semantic content. By precisely compressing dimensions within the V-L PTMs, RFE facilitates better alignment between the feature spaces of point clouds and their corresponding embeddings. On the other hand, SNC is a graph-based 3D model tailored to extract robust geometric information from point clouds. This additional layer of information enhances the geometric relationships between object components, mitigating the loss of knowledge that occurs during projection, especially in noisy, real-world scanned data. As a result, our FILP-3D framework gains improved resilience against noise interference.
Furthermore, we observe that the numbers of samples for different classes in test datasets can vary by up to two orders of magnitude, leading to a significant imbalance. This imbalance skews traditional accuracy metrics, making them less reliable for comprehensive evaluation. To address this, we introduce new evaluation metrics, namely $\rm NCAcc$ and $\rm F_{FSCIL}$. These metrics are designed to assess the model's ability to learn new classes effectively while maintaining a balance between old and new classes, thereby providing a more nuanced evaluation of the model's performance.

In a nutshell, our contributions are three-fold:
\begin{itemize}
        \item We pioneer the application of Vision-Language Pre-Training Models (V-L PTMs) to 3D Few-Shot Class-Incremental Learning (FSCIL) tasks, achieving performance gains over existing models. The general embedding provided by V-L PTM, along with its embedded prior knowledge, can complement the missing information in few-shot tasks, alleviate catastrophic forgetting, and narrow the domain gap during the incremental learning process.
        \item We introduce FILP-3D, a framework that incorporates two innovative modules: the Redundant Feature Eliminator (RFE) and the Spatial Noise Compensator (SNC). RFE specifically addresses feature space misalignment, while SNC is designed to mitigate the adverse effects of noise on the model. FILP-3D yields notable performance improvements, especially in metrics associated with novel classes.
        \item We develop new metrics, namely $\rm NCAcc$ and $\rm F_{FSCIL}$ to provide a more nuanced evaluation. These metrics assess a model's ability to adeptly learn new classes without compromising the performance on existing classes, offering a more comprehensive evaluation framework for 3D FSCIL tasks.
\end{itemize}

%% file: sections/2_related_work.tex
\section{Related Work}
\subsection{Vision-language Pre-training}
Vision-language (V-L) pre-training has become a popular topic in multi-modal tasks. Pre-trained by large-scale image-text pairs, those models are capable of many downstream tasks, \textit{e.g.}, visual question answering, and text-to-image generation. CLIP~\cite{CLIP} further leverages V-L pre-training to transfer cross-modal knowledge from the massive image-text pairs (400M), allowing the natural language to understand visual concepts, and exhibit strong capability in many tasks. ~\citet{zhu2023not, kang2024bibimbap} utilize CLIP to reduce the domain gap via CLIP’s prior knowledge in the domain adaption task. ~\citet{CLIP1, CLIP2} leverage general and adaptive embeddings of CLIP to prevent forgetting during the incremental learning stage. ~\cite{zhang2022pointclip, CLIP2Point, ULIP} verify that shape-related prior knowledge can enhance the model's few-shot/zero-shot capability even in 3D point cloud modal.

\textcolor{red}{Following CLIP, recent models like BLIP-2~\cite{li2023blip} incorporated iterative learning strategies and efficient cross-attention mechanisms to refine vision-language representations further. More recent work focuses on expanding these models' capabilities to handle video-language pre-training. For example, LLaVA~\cite{liu2024visual} integrates large language models with visual inputs to enable multimodal dialogue and image understanding, marking a step forward in interactive VLM applications. These models collectively push the boundary of generalization, robustness, and adaptability in vision-language tasks, \textit{e.g.}, open-world video recognition~\cite{CHEN2024111189}.}

\subsection{3D Point Cloud understanding}
In recent years, many works have been proposed to classify 3D point cloud objects. PointNet~\cite{PointNet} design the architecture to maintain natural invariances of the data. DGCNN~\cite{DGCNN} connects the point set into a graph and designs a local neighbor aggregation strategy. ~\citet{hao2023contrastive} propose CGRL to enhance 3D point cloud classification by generating more discriminative features, particularly in zero-shot scenarios. And some works employ masked point modeling as a 3D self-supervised learning strategy to achieve great success. For example, Point BERT~\cite{yu2022point} uses a pre-trained tokenizer to predict discrete point labels, while Point-MAE~\cite{pointmae} applies masked autoencoders to directly reconstruct the masked 3D coordinates.

Recently, inspired by the breakthroughs in V-L PTMs~\cite{CLIP}, a number of approaches have been suggested to transfer 2D PTM to point cloud tasks and show excellent performance. CLIP2Point~\cite{CLIP2Point} transfers CLIP to point cloud classification with image-depth pre-training. I2P-MAE~\cite{zhang2023learning} leverages knowledge of 2D PTMs to guide 3D MAE and ULIP~\cite{ULIP} improves 3D understanding by aligning features from images, texts, and point clouds.
These models are supported by prior knowledge from V-L PTMs, which can improve few-shot/zero-shot performance. 

Although these models have achieved remarkable progress in the 3D few-shot/zero-shot domain, these 3D models still exhibit limited generalizability, \textit{e.g.}, they can hardly handle OOD examples like real-scanned data. Incremental learning can more effectively mine knowledge from real-world scenarios, thus benefiting downstream applications. In this work, we aim to introduce V-L PTMs to the 3D FSCIL task and further improve the generalizability and performance.

\subsection{Few-Shot Class-Incremental Learning}
The FSCIL problem was proposed by~\citet{FSCIL}. Concretely, FSCIL aims at learning from severely insufficient samples incrementally while preserving already learned knowledge. TOPIC~\cite{FSCIL} uses a neural gas network to learn and preserve the topology of features. Subsequently, PFR~\cite{pan2024pseudo} employs a pseudo-set training strategy with frequency-based feature refinement to mitigate the lack of discriminative feature learning. Also, some models like FACT~\cite{FACT} try to use virtual prototypes to reserve for new ones, thus ensuring incremental learning ability. The SOTA method BiDist~\cite{BiDist} utilizes a novel distillation structure to alleviate the effects of forgetting. 

Recently, with the breakthroughs in 2D PTMs, a number of works have attempted to leverage the vast knowledge acquired by 2D PTMs, which is effective in learning new concepts and alleviating the problem of forgetting~\cite{CLIP1, CLIP2}. With almost perfect performance, these models have attracted a lot of interest and attention.

The preceding methods are all 2D methods, except for \citet{FSCIL-3D}'s work which explores the FSCIL task in 3D point cloud data. \citet{FSCIL-3D} points out the domain gap between synthetic instances from the base task and real-scanned samples in incremental steps is a specific issue of 3D FSCIL. The domain gap exacerbates the inherent issues of catastrophic forgetting and overfitting in the FSCIL task. To address this, they propose microshapes to minimize variation between synthetic and real data and achieve domain adaptation. 

However, \citet{FSCIL-3D}'s work neither supplements knowledge for few-shot data and domain gap nor addresses the high-noise nature of real-world scanned data, resulting in limited performance in 3D FSCIL tasks. In contrast, our FILP-3D successfully addresses the shortcomings mentioned above by introducing V-L PTMs and two newly proposed modules.

%% file: sections/2.5_symbol_table.tex
\section{Notations Table}
\textcolor{red}{In this section we first provide the notation of symbols involved in Sec.~\ref{sec:method} and Sec.~\ref{sec:benchmark}, as shown in Tab.~\ref{tab:notation}. Note that in Sec.~\ref{sec:method}, $\mathcal{D}$ refers to the training datasets, whereas in Sec.~\ref{sec:benchmark}, $\mathcal{D}$ refers to the testing datasets. And in Sec.~\ref{sec:method} (except Section~\ref{sec:RFE} Details of Pre-Processing), we replace $\mathbf{x}_{i}^{b}$, $y_{i}^{b}$ with $\mathbf{x}$, $y$ for clearer explanation.}
\begin{table}[!t]
    \centering
    \caption{Notations table.}
    \label{tab:notation}
    \fontsize{8}{9}\selectfont
    \setlength{\tabcolsep}{15pt}
    \begin{tabular}{l l}
    \toprule
    \multicolumn{2}{l}{\textbf{Symbols in Dataset}} \\
    \midrule
    $B$ & the number of task \\
    $P$ & the number of points contained in a point cloud \\
    $\mathcal{D}$ & the set of dataset in each session\\
    $\mathcal{D}^b$ & the dataset of the $b$-th session\\
    $n_b$ & the number of point cloud in $\mathcal{D}^b$ \\
    $\mathbf{x}_{i}^{b}$ & the $i$-th point cloud in $\mathcal{D}^b$  \\
    $y_{i}^{b}$ & the label of $i$-th point cloud in $\mathcal{D}^b$  \\
    $Y_{b}$ & the label set of the $b$-th session\\ 
    $K_b$ & the number of visible classes for the $b$-th task \\
    \midrule
    \multicolumn{2}{l}{\textbf{Symbols in Method}} \\
    \midrule
    $N$ & the number of views in the multi-view render stage\\
    $C$ & the embedding dimension of V-L PTM\\
    $M$ & the number of principal components\\
    $K$ & the number of the visual classes in this session\\
    $\mathbf{f}^p$ & the raw 3D feature obtained from the input point cloud $\mathbf{x}$ via the 3D encoder\\
    $\mathbf{f}^{3D}$ & the 3D feature modified from the raw 3D features $\mathbf{f}^{3D}$ via the adapter \\
    ${\rm \mathbf{D}}$ & the multi-view depth maps rendered from the input point cloud $\mathbf{x}$\\
    ${\rm \mathbf{F}}^{d}$ & the multi-view depth features obtained from the multi-view depth maps ${\rm \mathbf{D}}$ via the V-L PTM\\
    $\mathbf{f}^{2D}$ & the global 2D feature obtained from the multi-view depth maps ${\rm \mathbf{F}}^{d}$ via the Merger \\
    $t_k$ & the label name of $k$-th class\\
    ${\rm \mathbf{F}}^t$ & the raw text features obtained from visible label set via V-L PTM and the template text
prompt:\\
    $\mathbf{f}^g$ & the global feature fused from the 3D features $\mathbf{f}^{3D}$ and the global 2D feature $\mathbf{f}^{2D}$ via the fusion fuction\\
    $\mathbf{V}$ & the pre-processed principal components used to eliminate redundant dimensions\\
    $\widetilde{\mathbf{f}}^g$ & the global feature with redundant dimensions eliminated\\
    $\widetilde{\mathbf{F}}^t$ & the text features with redundant dimensions eliminated\\
    $l_k$ & the logit for class $k$-th class \\
    $\mathbf{p}$ & the ultimate prediction probability \\
    $N_{Aug}$ & the number of augmentation iterations during the contrastive learning stage\\
    \midrule
    \multicolumn{2}{l}{\textbf{Symbols in Benchmark}} \\
    \midrule
    ${\rm Acc_i}$ & the accuracy of $i$-th session.\\
    ${\Delta}$ & relative accuracy dropping rate, where ${\rm \Delta=\frac{|Acc_B-Acc_1|}{Acc_1}}$ \\
    $Acc_i$ & the accuracy of the $i$-th class \\
    \bottomrule
    \end{tabular}
\end{table}

%% file: sections/3_method.tex
\section{Proposed Method}\label{sec:method}
\subsection{Problem Formulation}
Assuming a sequence of $B$ tasks $\mathcal{D} = \{\mathcal{D}^1,\mathcal{D}^2,\dots,\mathcal{D}^B\}$, FSCIL methods incrementally recognize novel classes with a small amount of training data. For the $b$-th task $\mathcal{D}^b = \{(\mathbf{x}_{i}^{b},y_{i}^{b})\}_{i=1}^{n_b}$, we have $n_b$ training samples, in which an instance $\mathbf{x}_{i}^{b}$ has a class label $y_i^b$ and $y_i^b$ belongs to the label set $Y_b$. We stipulate that $Y_{b} \cap Y_{b'} = \emptyset$ when $b \not= b'$.
After the $b$-th training task, trained models are required to classify test sets of all the previous tasks $\{\mathcal{D}^{1}, \dots, \mathcal{D}^{b}\}$.
Note that, in the 3D FSCIL setting, $\mathbf{x}_{i}^{b}\in\mathbb{R}^{P\times 3}$ denotes a point cloud object, where $P$ denotes the number of points in a point cloud object. $\mathcal{D}^1$ is the base task with a large-scale 3D training dataset, while much fewer samples are included in the incremental task $\mathcal{D}^b$, \textit{i.e.}, $n_b \ll n_1$ for $b > 1$.
\subsection{Model Overview}\label{sec:overview}

\begin{figure*}[t]
  \centering
  \includegraphics[width=1.0\textwidth]{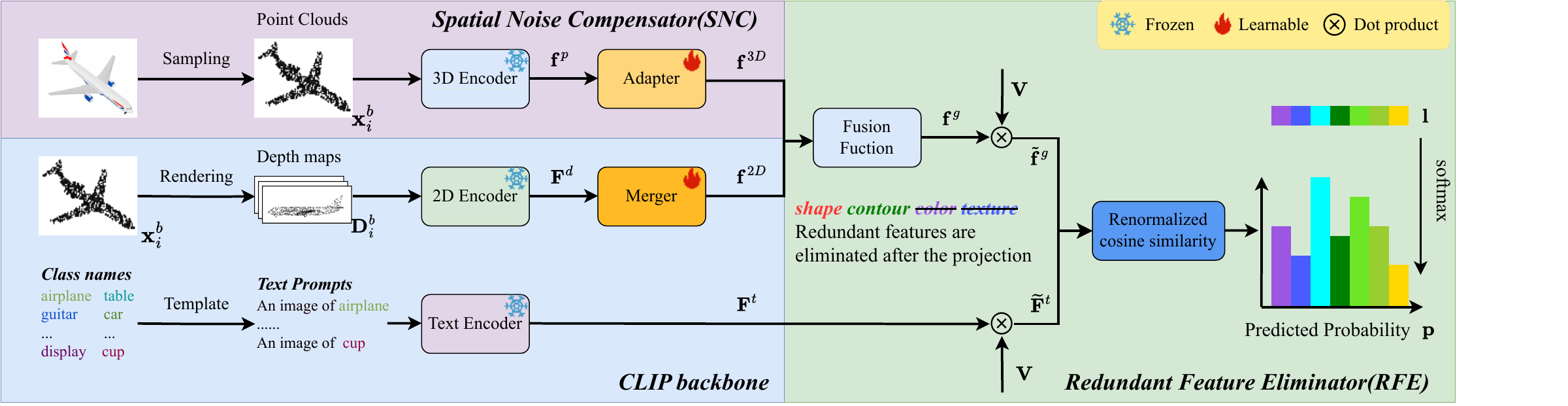}
  \caption{\textcolor{red}{Overview of FILP-3D. FILP-3D mainly consists of three components, \textit{i.e.}, 3D branch (SNC), CLIP backbone, and the classify component (RFE). The SNC generates 3D feature $\mathbf{f}^{3D}$ from the input point cloud. The CLIP backbone also generates global 2D feature $\mathbf{f}^{2D}$ and text features ${\rm \mathbf{F}}^t$ from the input point cloud and the class names respectively. Then, the 3D feature $\mathbf{f}^{3D}$ and the global 2D feature $\mathbf{f}^{2D}$ will be fused as a global feature $\mathbf{f}^g$, and be used to calculate probability alongside text features after redundant dimensions eliminated in the RFE module.}}
  \label{fig: overview}
  \vspace{-1.0em}
\end{figure*}
The framework of FILP-3D is shown in Figure~\ref{fig: overview}.
For each training sample $(\mathbf{x},y)$, we mainly have three branches: 1) point cloud $\mathbf{x}$ is fed into 3D encoder to obtain an original point feature $\mathbf{f}^{p}$ and then adapted to the 3D feature $\mathbf{f}^{3D}$, 2) $\mathbf{x}$ is then rendered as multi-view depth maps ${\rm \mathbf{D}}$, embedded as multi-view depth features ${\rm \mathbf{F}}^{d}$, and finally merged into a global 2D feature $\mathbf{f}^{2D}$, 3) visible class names are templated into text prompts and encoded as text features ${\rm \mathbf{F}}^t$.
The 3D features $\mathbf{f}^{3D}$ and the global 2D feature $\mathbf{f}^{2D}$ are then fused as a global feature $\mathbf{f}^g$.
Afterward, we use pre-processed principal components $\mathbf{V}$ (details of pre-processing will be mentioned in Sec. 4.4.1) to eliminate redundant dimensions of $\mathbf{f}^g$ and $\mathbf{F}^t$ and generate $\widetilde{\mathbf{f}}^g$ and $\widetilde{\mathbf{F}}^t$. The final logits $l$ are calculated by our proposed renormalized cosine similarity and the predicted probability $\mathbf{p}$ is the result of applying the softmax function to the logits $l$.


\subsection{PTM is a Good 3D FSCIL Learner}
\textcolor{red}{Recent works~\cite{CLIP1, CLIP2} have shown that PTMs substantially enhance performance in incremental learning tasks. However, \citet{FSCIL-3D}'s method lacks shape-related prior knowledge and naturally aligned features/text encoders, thus struggling to retain old knowledge while learning new knowledge.}

\textcolor{red}{In response to the aforementioned challenges, we attempt to introduce PTM to incorporate prior knowledge.} Our initial approach was to embed shape-related prior knowledge via a 3D PTM~\cite{yu2022point}. However, the substantial disparity in quality and volume between 3D data~\cite{shapenet,co3d} and its text/2D counterparts~\cite{CLIP} renders current 3D PTMs less effective in generalizing to downstream tasks, like zero/few-shot classification~\cite{CLIP2Point}. Motivated by this limitation, we turn to an alternative method. Drawing inspiration from recent studies~\cite{zhang2022pointclip, CLIP2Point}, we leverage CLIP~\cite{CLIP} to indirectly infuse shape-related prior knowledge through projection. We christen this new framework as our baseline.

Specifically, we project the point cloud data into multi-view depth maps. A pre-trained ViT~\cite{ViT} in CLIP is then deployed to extract depth features $\mathbf{F}^d\in\mathbb{R}^{N\times C}$. Here, $N$ is the number of views, and $C$ is the embedding dimension of ViT. To allow incremental tasks, a learnable merger is attached, formulating global 2D features $\mathbf{f}^{2D}\in\mathbb{R}^C$ as follows,
\begin{equation}\label{eq1}
    \mathbf{f}^{2D} = {\rm E}(\Phi({\rm F}^d))
\end{equation}
where $\rm E$ is a two-layer perceptron with ReLU activation function and $\Phi$ denotes the concat operation. This simple adapter can effectively act as a \textbf{Merger} to merge multi-view features into a global feature~\cite{CLIP2Point}.

The efficacy of setting classifier weights to average embeddings (referred to as prototypes) for CIL tasks has been well-established by~\cite{prototype}. However, while \citet{FSCIL-3D}'s method employs word2vec to construct prototypes for new classes, it falls short in encapsulating the average semantics of 3D point clouds. In our approach, the 2D features are pre-aligned with text embeddings of CLIP. This enables our baseline to progressively produce prototypes using a CLIP text encoder. For each class labeled as $k$ with its respective name $t_k$, we introduce a template text prompt: ``\textit{an image or projection or sketch of a} $t_k$". This is mapped to the CLIP prototype symbolized by $\mathbf{F}^t_k\in\mathbb{R}^C$. Owing to the inherent association of our method between image and textual representations, there is no compulsion to realign the two modalities during incremental phases, unlike strategies such as in \citet{FSCIL-3D}'s work. As a result, we can directly utilize the cosine similarity between $\mathbf{f}^{2D}$ and $\mathbf{F}^t_{k}$ to represent the logit for class $k$. The ultimate probability prediction, represented by $\mathbf{p}$, is formulated as follows:
\begin{equation}\label{eq2}
    l_k =  \cos(\mathbf{f}^{2D}, \mathbf{F}^{t}_k), \quad \mathbf{p} = {\rm softmax}([l_1, \dots, l_{K}]).
\end{equation}
where $\cos(\cdot,\cdot)$ denotes the cosine similarity, and $K$ is the number of visual classes.

Despite these improvements, the incompatibility between 3D data representation and the 2D feature encoder impede further performance enhancements, primarily manifested as two aspects.

\subsection{Ensure Performance of PTM in 3D FSCIL}
\textcolor{red}{Large-scale V-L PTMs undeniably serve as rich repositories of prior knowledge. However, their application to 3D FSCIL tasks reveals unique challenge: mismatch between mismatch between 3D representation and V-L PTM’s encoder.
This mismatch is primarily manifested in two aspects: feature space misalignment and significant noise. In the ensuing sections, we aim to enhance the baseline with elucidating solutions tailored to these specific hurdles, to obtain \textbf{FILP-3D}. Specifically, we propose the Redundant Feature Eliminator (RFE) to address the feature space misalignment issue and the Spatial Noise Compensato (SNC) to tackle the problem of significant noise.}

\subsubsection{Redundant Feature Eliminator}~\label{sec:RFE}
We propose our design to mitigate the adverse effect of feature space misalignment here. Our empirical observations indicate that redundant features typically exhibit minimal inter-class distinction, leading to constrained variance. By contrast, semantic features pertinent for classification distinctly express a pronounced bias for each class, resulting in a more pronounced variance. ~\citet{FSCIL-3D,zhu2023not} also yield empirical observations akin to ours. Drawing upon this insight, we can discern between redundant and semantic features based on their variance. Subsequently, we employ Dimensionality Reduction (DR) techniques to preserve the semantic essence while condensing the extraneous features. 

\begin{figure}[t]
  \centering
  \includegraphics[width=1.0\textwidth]{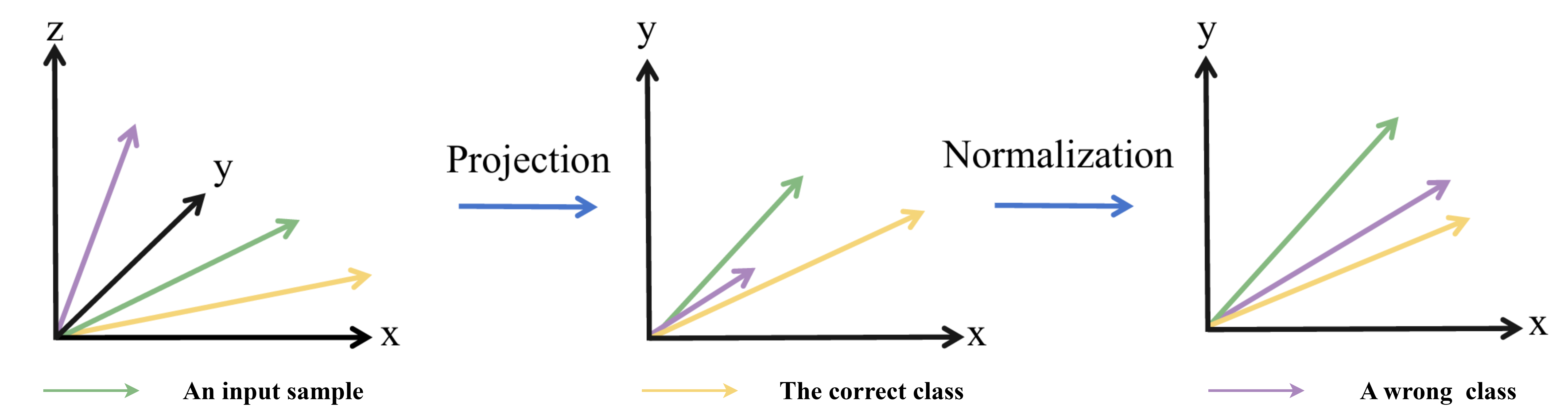}
  \caption{The current feature space is generated by three principal components. Each dimension's one-hot vector represents a principal component. \textcolor{red}{The first two dimensions contain semantic information, while the third dimension serves as a redundant component. By transforming feature vector (green) into the feature space mentioned above, we can notice that projection can eliminate redundant information, while normalization will improperly stretch the semantic information, leading to misclassification.}}
  \label{fig: sample}
  \vspace{-1.0em}
\end{figure}

Following the above discussions, we can obtain the formulation for calculating $\widetilde{l}_k$, $i.e.$, the logit after eliminating superfluous features. The detailed discussion is as follows.

$\mathbf{V}$ is the principal components extracted from the base task, $\mathbf{v}_i$ is the $i$-th normalized principal component. $\widetilde{\mathbf{f}}^{2D}$ and $\widetilde{\mathbf{F}}^t_k$ are the projection of $\mathbf{f}^{2D}$ and $\mathbf{F}^t_k$ onto the $\mathbf{V}$ vector space, formally expressed as the following equation:
\begin{gather}\label{Eq3}
    \mathbf{f}^{2D} = \sum_{i=1}^{M}\widetilde{\mathbf{f}}^{2D}_i\mathbf{v}_i + \mathbf{R}^{2D}\ 
    \mathbf{F}^{t}_k = \sum_{i=1}^{M}\widetilde{\mathbf{F}}^t_{k,i}\mathbf{v}_i + \mathbf{R}^t_{k}
\end{gather}
where $M$ is the number of principal components, $\widetilde{\mathbf{f}}^{2D}_i$ and $\mathbf{F}^t_{k,i}$ are the $i$-th dimension of $\widetilde{\mathbf{f}}^{2D}$ and $\widetilde{\mathbf{F}}^t_k$, respectively, and $\mathbf{R}^{2D}$ and $\mathbf{R}^t_{k}$ are the components of $\mathbf{f}^{2D}$ and $\mathbf{F}^{t}_k$ which are not in the $\mathbf{V}$ vector space. 
$\mathbf{R}^{2D}$ and $\mathbf{R}^t_{k}$ are \textbf{low-variance components} (due to the nature of principal component analysis) and thus are \textbf{redundant components} according to previous discussions.

Each $\mathbf{v}_i$ is an orthogonal vector and orthogonal to vectors outside the $\mathbf{V}$ vector space, e.g., $\mathbf{R}^{2D}_k$ and $\mathbf{R}^t_{k}$. Therefore $\forall i,j,k,\ i \neq j$ The following equation holds:
\begin{gather}\label{Eq1}
    \mathbf{v}_i\mathbf{v}^T_j = 0,\ \mathbf{v}_i\mathbf{v}^T_i = 1,\ 
    \mathbf{R}^{2D}_k\mathbf{v}_i^T = 0,\ \mathbf{v}_i\mathbf{R}^{t^T}_k = 0 
\end{gather}
Then, we can derive the following equation:
\begin{align*}
  l_k = \frac{\mathbf{f}^{2D} \mathbf{F}^{t^\mathrm{T}}_k}{\|\mathbf{f}^{2D}\| \|\mathbf{F}^t_k\|} 
  &= \frac{(\sum_{i=1}^{M}\widetilde{\mathbf{f}}^{2D}_i\mathbf{v}_i + \mathbf{R}^{2D})(\sum_{i=1}^{M}\widetilde{\mathbf{F}}^t_{k,i}\mathbf{v}_i + \mathbf{R}^t_{k})^T} {\|\mathbf{f}^{2D}\| \|\mathbf{F}^t_k\|} \\
  &= \frac{\widetilde{\mathbf{f}}^{2D}\widetilde{\mathbf{F}}^{t^T}_k + \mathbf{R}^{2D}\mathbf{R}^{t^{T}}_k}{\|\mathbf{f}^{2D}\|\|\mathbf{F}^{t}_k\|}
\end{align*}

Now we can observe that there are two terms in the numerator of this equation. $\widetilde{\mathbf{f}}^{2D}\widetilde{\mathbf{F}}^{t^T}_k$ represents semantic information and should be retained, and $\mathbf{R}^{2D}\mathbf{R}^{t^T}_k$ is the multiplication of redundant feature vectors that should be compressed to keep only semantic information. Thus, the logit of the class $k$ can be calculated in a modified way as follows to eliminate redundant features:
\begin{gather}\label{eq5}
    \widetilde{l}_k = \frac{\widetilde{\mathbf{f}}^{2D}\widetilde{\mathbf{F}}^{t^T}_k}{\|\mathbf{f}^{2D}\|\|\mathbf{F}^{t}_k\|}.
\end{gather}

Note that, different from calculating the cosine similarity of $\widetilde{\mathbf{f}}^{2D}$ and $\widetilde{\mathbf{F}}^t_k$, the \emph{denominator} in Eq.~\ref{eq5} is the norm of the original features $\mathbf{f}^{2D}$ and $\mathbf{F}^t_k$.
Principal features are directly dot multiplied after normalizing with the original features, and such a procedure is named as the \textbf{renormalized cosine similarity (RCS)}. 
\textcolor{red}{As shown in Figure~\ref{fig: sample}. Both the direction and length of the projected vectors influence the classification result in RCS. However, the regularization operation in direct cosine similarity disregards the projected vector's length, which leads to incorrect classification. For example: The class of the input sample is airplane and the wrong class is table. The x and y axes denote the presence or absence of wings and the shape of the backbone. The z-axis denote the surface texture. The airplane has wings, an oval backbone, and metal surface, so the yellow and green arrows are high in the x and y axes and low in the z-axis. The table does no have a wing and has a rectangle backbone and wooden surface, so the purple arrow is low in the x and y axes and high in the z-axis. After projection, the yellow and green arrows are high along both the x and y axes, while the purple arrow is low along both axes, allowing the sample to be correctly classified as an airplane. However, after normalization, the purple arrow becomes high along both the x and y axes as well, causing the sample to be misclassified as a table.}

\noindent\textbf{Details of Pre-Processing.}
\begin{figure}[t]
  \centering
  \includegraphics[width=0.8\textwidth]{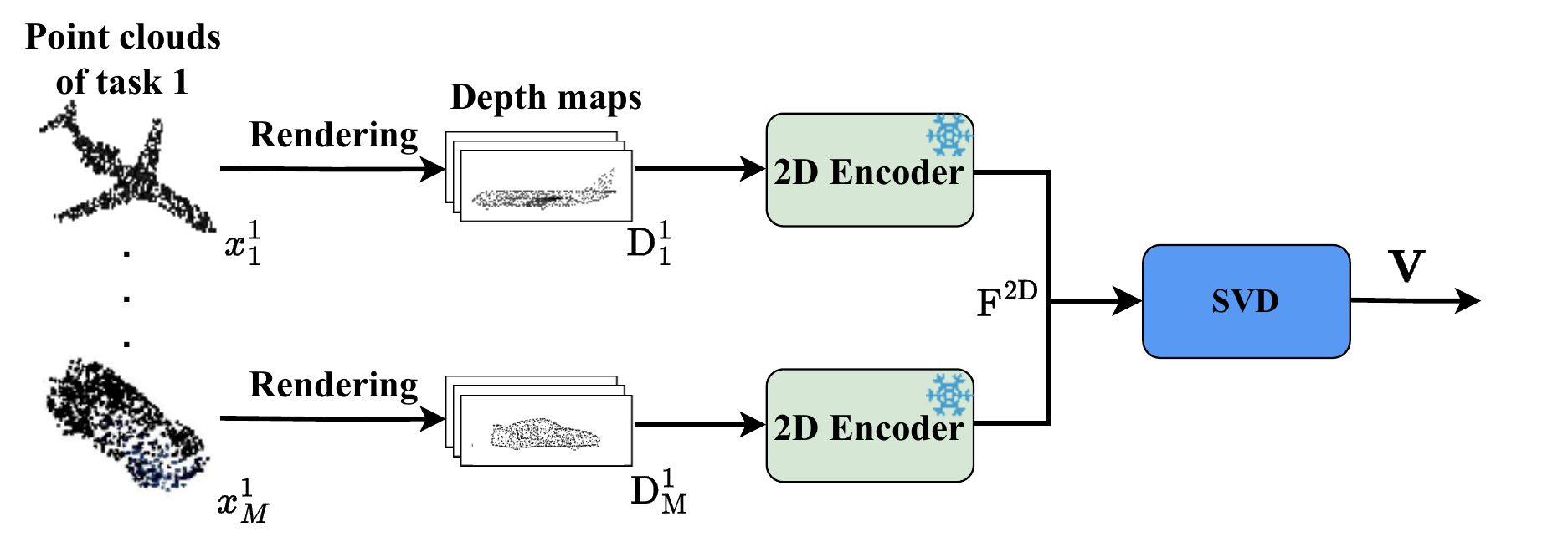}
  \caption{Overview of pre-processing}
  \label{fig: Pre-Processing}
\end{figure}
In the pre-processing stage, we render all training samples of the base task $\{\mathbf{x}^1_1,...,\mathbf{x}^1_M\}$ to generate their depth maps $\{\rm \mathbf{D}^1_1,...,\mathbf{D}^1_M\}$ and then embed them as depth features $\rm \mathbf{F}^{d}\in\mathbb{R}^{(|\mathcal{D}^1|N) \times C}$ by the ViT in CLIP, where $|\mathcal{D}^1|$ is the number of training sample in task 1, $N$ is the number of viewpoint and $C$ is the embedding dimension. We then use the depth features ${\rm \mathbf{F}^{2D}}$ to calculate principal components $\mathbf{V}\in\mathbb{R}^{M \times C}$ through SVD, as is shown in Figure~\ref{fig: Pre-Processing}, where $M$ is the number of principal components.

\subsubsection{Spatial Noise Compensator}
Baseline unifies pre-aligned image and text encoders through comprehensive data training. This ensures a substantial reservoir of prior knowledge, minimizing the potential for information loss during any subsequent alignment processes. Given these attributes, it stands as a markedly superior choice when juxtaposed against partially aligned 3D pre-trained models. Nonetheless, the projection-based methodology inherent to the baseline exhibits heightened sensitivity to both noise and viewpoint selection. When noise obscures the clarity of depth maps, rendering them ineffective in delineating object contours, there is a marked downturn in classification efficacy. The supervised classification performance of other projection-based methods like PointCLIP~\cite{zhang2022pointclip} (72.8\% Acc.) and CLIP2Point~\cite{CLIP2Point} (74.5\%) compared to DGCNN~\cite{DGCNN} (78.1\%) and PointMAE~\cite{pointmae} (86.4\%) on ScanObjectNN also verify ours viewpoint. As a countermeasure, we propose the incorporation of a graph-based 3D model. This addition is devised to augment the compromised information, bolstering our model's resilience against noise disturbances.

The 3D module yields features that are invariant to transformations, thereby enhancing their robustness against noise. On the other hand, the multi-view model, fortified by PTM, extracts semantically richer and more encompassing information. To harness the strengths of both, we amalgamate the features derived from the 3D module and the multi-view model. 

Following ULIP~\cite{ULIP}, FILP-3D adapts cross-modal contrastive learning to achieve alignment between 3D and CLIP text/image.
This alignment operation maps the features of 3D, image, and text to the same feature space, enabling a more rational fusion of 2D and 3D features and effective classification.
The align loss $\mathcal{L}_{(M1,M2)}$ between modal $M1, M2$ and the final align loss $\mathcal{L}_{align}$ can be calculated as follows:
\begin{gather}
    \label{contloss}
    \mathcal{L}_{(M1,M2)}=\sum_{(i,j)}-\frac12log\frac{\exp\left(\frac{\mathbf{h}_{i}^{M_{1}}\mathbf{h}_{j}^{M_{2}}}{\tau}\right)}{\sum_{k}\exp\left(\frac{\mathbf{h}_{i}^{M_{1}}\mathbf{h}_{k}^{M_{2}}}{\tau}\right)}  
    -\frac{1}{2}log\frac{\exp\left(\frac{\mathbf{h}_i^{M_1}\mathbf{h}_j^{M_2}}{\tau}\right)}{\sum_k\exp\left(\frac{\mathbf{h}_k^{M_1}\mathbf{h}_j^{M_2}}{\tau}\right)}\\
    \label{align}
    \mathcal{L}_{align} = \alpha\mathcal{L}_{(P,S)} + \beta\mathcal{L}_{(I,S)}
\end{gather}
where $\mathbf{h}_{i}^{M}$ donates the $i$-th feature in modal $M$, $\tau$ donates the temperature parameter and the $P, I, S$ donate the set of point cloud modal $\mathbf{f}^{3D}$, the set of image modal $\mathbf{f}^{2D}$ and text modal $\mathbf{F}^{t}$ in this session respectively. $\mathbf{f}^{3D}$ can be calculated as follows:
\begin{gather}
    \label{eq7}
    \mathbf{f}^{3D} = {\rm E}({\rm \mathbf{f}^p})
\end{gather}
where $\rm E$ is a two-layer perceptron with ReLU activation function and is named as the \textbf{Adapter}.

After the alignment, we utilize the mixed pooling~\cite{yu2014mixed} to fuse 2D features and 3D features. According to~\cite{yu2014mixed}, mixed pooling is superior to single max/average pooling in terms of feature fusion. Using this approach, any information inadequately captured due to viewpoint selection is supplemented by the 3D channel using the ${\rm max}$ operation. Concurrently, overly dominant features are moderated through the ${\rm average}$ operation for balance. The new global feature can be calculated as follows:
\begin{gather}
    \label{eq8}
    \mathbf{f}^g=\frac{1}{2}({\rm max}(\mathbf{f}^{2D},\mathbf{f}^{3D})+{\rm average}(\mathbf{f}^{2D},\mathbf{f}^{3D}))
\end{gather}
where $\mathbf{f}^g$ represents global features used to replace $\mathbf{f}^{2D}$ for further tasks. 


\subsection{Training Objective}
The parameters of the 3D encoder and the multi-view render \& encoder are fixed during both base and incremental training stages. Only Merger (Eq.~\ref{eq1}) for depth feature fusion and Adapter (Eq.~\ref{eq7}) for 3D feature alignment are trained. The classification loss $\mathcal{L}^b_{cls}$ for $b$-th task can be calculated as follows:
\begin{equation}
    \label{loss1}
    \mathcal{L}^b_{cls} = \frac{1}{|\mathcal{D}^b|}\sum_{i=1}^{|\mathcal{D}^b|}\mathcal{L}_{ce}(\mathbf{p}^b_i,y^b_i)
\end{equation}
where $\mathcal{L}_{ce}$ is cross-entropy loss~\cite{ce}, $\mathbf{p}^b_i$ denotes the predicted probability of the $i$-th sample of the $b$-th task.

Inspired by ~\cite{song2023learning}, we also utilize contrastive learning~\cite{caron2020unsupervised} and data augmentation in the training of 3D FSCIL models for continual learning. 
Through the push and pull dynamics of contrastive learning, we optimize the extracted features, bringing them closer to the appropriate prototype while distancing them from erroneous ones. This strategy notably reduces ambiguity throughout the continual learning phase. Concurrently, data augmentation not only amplifies the efficacy of contrastive learning but also acts as a safeguard against overfitting in few-shot scenarios.

Specifically, we employ random rotations along the coordinate axes and random variations in the camera view distance as the augmentation function $f_{Aug}$. For the $i$-th sample of the $b$-th task, we train using the corresponding prototype $\mathbf{F}^t_{y^b_i}$ as a positive example and all other visible prototypes $\mathbf{F}^t_{res^b_i}$ as negative examples. We employ the InfoNCE loss~\cite{oord2018representation} as contrastive learning loss. The contrastive learning loss $\mathcal{L}^b_{cont}$ for the $b$-th task can be calculated as follows:
\begin{equation}
    \label{loss2}
    \mathcal{L}^b_{cont} = \frac{1}{|\mathcal{D}^b|}\sum_{i=1}^{|\mathcal{D}^b|}\sum_{j=1}^{N_{Aug}}\mathcal{L}_{InfoNCE}(\mathbf{f}^g_{i,j},\mathbf{F}^t_{y^b_i}, \mathbf{F}^t_{res^b_i})
\end{equation}
where $N_{Aug}$ is the number of augmentation iterations, and $\mathbf{f}^g_{i,j}$ is the global feature encoded after replacing $x^b_i$ with $f_{Aug}(x^b_i)$.

The overall training loss $\mathcal{L}^b$ for $b$-th task can be calculated as the sum of align loss Eq~\ref{align}, classification loss\ref{loss1} and contrastive learning loss~\ref{loss2}  follows:
\begin{equation}
    \label{loss3}
    \mathcal{L}^b = \mathcal{L}^b_{align} + \mathcal{L}^b_{cls} + \mathcal{L}^b_{cont}
\end{equation}
\section{Benckmark for 3D FSCIL task}\label{sec:benchmark}
Studies on 3D FSCIL benchmarks are still in the early stages. The sole benchmark introduced by \citet{FSCIL-3D} has several notable limitations:
1) In the synthetic data to synthetic data (\textbf{S2S}) task set forth by \citet{FSCIL-3D}, the total number of classes is limited. This restricts the capacity of the model for incremental learning and makes it challenging to evaluate its effectiveness comprehensively.
2) For the synthetic data to real-scanned data \textbf{(S2R}) task, many analogous classes have been removed without clear justification.
3) Relying solely on accuracy as a metric means that incremental classes may not receive the emphasis they warrant.
In light of these observations, we introduce a new benchmark: FSCIL3D-XL. Details of this benchmark will be discussed in the subsequent sections.

\subsection{Task Setting}
FSCIL3D-XL comprises two series of incremental tasks: S2S and S2R.
The S2S task acts as a transitional and simulation-based task, primarily designed to assess the model's capability to mitigate issues like overfitting or catastrophic forgetting. On the other hand, the S2R task is inherently more complex and has broader applicability. Given its larger domain gap and challenges with noise, it demands a more robust ability to generalize from limited samples.

\noindent\textbf{S2S Task.}
\citet{FSCIL-3D}'s dataset is constructed using a single dataset, encompassing only 55 classes. In our benchmark, we opt for ShapeNet~\cite{shapenet} as our base dataset and ModelNet40~\cite{modelnet} as the incremental dataset.
Our base task retains all the 55 classes from ShapeNet~\cite{shapenet}. For the incremental tasks, we exclude 16 classes from ModelNet40~\cite{modelnet} that overlap with classes in the base task, and the remaining 24 unique classes from ModelNet40 are then evenly distributed across 6 incremental tasks. In contrast to \citet{FSCIL-3D}'s work, our S2S task offers an increase in class count by over 40\%, introducing a heightened level of challenge.


\noindent\textbf{S2R Task.}
\citet{FSCIL-3D}'s benchmark unjustifiably excludes certain classes that, while similar, are distinct in fact. For instance, semantically similar classes like ``handbag" and ``bag", or shape-analogous classes like ``ashcan" and ``can", are filtered out. By contrast, we retain all available classes in our benchmark. Recognizing the subtle differences between such similar classes is undoubtedly challenging, but we believe it is essential for a comprehensive evaluation. We continue to use ShapeNet as the base dataset and select CO3D~\cite{co3d} as the incremental dataset. From CO3D, we exclude 9 overlapping classes, resulting in 41 distinct classes designated for the incremental tasks.

\subsection{Evaluation metrics}
3D datasets present challenges distinct from their 2D counterparts. Firstly, there is a pronounced imbalance in the scale of different classes within 3D datasets. For instance, in ShapeNet, the ``chair" class boasts over 1,000 training samples, whereas ``birdhouse" has a mere 15. Secondly, incremental samples in 3D datasets, particularly those sourced from real-scanned data, tend to be more intricate. These novel classes, given their complexity, demand heightened attention in the FSCIL task. Relying solely on the micro accuracy metric, denoted as ${\rm Acc}$, would inadequately address these challenges. \textcolor{red}{As a result, while we retain the 2D evaluation metrics ${\rm Acc_i}$ (micro accuracy of $i$-th session, where ${\rm Acc_i} = \rm{\frac{CP}{TP}}$. $\rm{CP}$ and $\rm{TP}$ denote the correct/total predictions in $i$-th session respectively.)} and ${\Delta}$~\cite{tan2022graph} (relative accuracy dropping rate, where ${\rm \Delta=\frac{|Acc_B-Acc_1|}{Acc_1}}$), we also introduce new evaluation metrics specifically designed to address the aforementioned issues.

\noindent\textbf{Macro accuracy.}
Macro accuracy (${\rm MAcc}$) indicates the generalization ability of the model, preventing overfitting in a small number of classes with a relatively large number of samples. It can be calculated using the following formula for $b$-th task:
\begin{equation}\label{MAcc}
{\rm MAcc_b}=\frac{1}{K_b}\sum_{i=1}^{K_b} Acc_i
\end{equation}
where $K_b=\sum_{i=1}^{b}|Y_b|$ denotes the number of visible classes for the $b$-th task, and $Acc_i$ denotes the accuracy of the $i$-th class.

\noindent\textbf{Novel class accuracy.}
Novel class accuracy (${\rm NCAcc}$) indicates the ability of the model to learn new classes, preventing the model from over-focusing on base classes with a large number of samples. It can be calculated using the following formula for $b$-th task:
\begin{gather}
\label{NCAcc1}
{\rm NCAcc_b}=\frac{|\{p_{i}^{b}|p_{i}^{b}=y_{i}^{b},p_{i}^{b}\in P_{b},y_{i}^{b}\in Y_{b}\}|}{|\mathcal{D}^b|} \\
\label{NCAcc2}
{\rm NCAcc}=\frac{1}{B}\sum_{b=1}^{B}{\rm NCAcc_b}
\end{gather}
where $P_b$ denotes the predicted labels for test stage of $b$-th task and $B$ denotes the number of task.
	
\noindent\textbf{F-Score.}
F-Score for FSCIL task (${\rm F_{FSCIL}}$, ${\rm F}$ for short): The network needs to be plastic to learn new knowledge from the current task, and it also needs to be stable to maintain knowledge learned from previous tasks. To ensure that the model is not overly biased towards either of these aspects, we refer to ${\rm F_{score}}$ and propose ${\rm F_{FSCIL}}$, which balances the plasticity and stability of the evaluation network. It can be calculated using the following formula:
\begin{equation}\label{F_FSCIL}
{\rm F_{FSCIL}}=\frac{2\ {\rm Acc_B}\ {\rm NCAcc}}{{\rm Acc_B} + {\rm NCAcc}}
\end{equation}
where ${\rm Acc_B}$ denotes the accuracy of the ${\rm B}$-th (final) session.

Finally, the 3D dataset is still being refined, so a large number of new datasets will likely continue to be produced. We take this into account and design our benchmark to be more flexible and modular in adding datasets, \textit{i.e.}, the desired FSCIL dataset can be obtained by simply transmitting the parameters and datasets to our generator. Additionally, The benchmark has been \href{https://github.com/HIT-leaderone/FILP-3D}{open-resource}.

%% file: sections/4_experiment.tex
\section{Experiments}
In this section, we begin by furnishing a more comprehensive exposition of our models and experimental setup. Subsequently, we proceed to present and analyze the results of the comparative experiments. Finally, we showcase ablation experiments and visualizations to substantiate the effectiveness of our proposed modules. Please note that we only utilize SNC during incremental learning with real-scanned data, such as in the S2R task. In cases where SNC is not applied such as the S2S task, we employ cosine similarity between $\mathbf{f}^{2D}$ and ${\rm \mathbf{F}}^t$ for classification.

\begin{table*}[!t]
  \caption{The number of samples $n_b$ in the training and testing sets for the S2S and S2R tasks}
  \centering
  \resizebox{0.8\linewidth}{!}{
  \begin{tabular}{lccccccccccccc}
        \toprule
        \multicolumn{1}{l}{\multirow{2}{*}{Task}} & \multicolumn{1}{l}{\multirow{2}{*}{Type}} & \multicolumn{12}{c}{The number of samples $n_b$}   \\
        \cmidrule{3-14}
        \multicolumn{1}{l}{} & \multicolumn{1}{l}{} & 0 & 1 & 2 & 3 & 4 & 5 & 6 & 7 & 8 & 9 & 10 & 11 \\
        \midrule
        \multicolumn{1}{l}{\multirow{2}{*}{ShapeNet2CO3D.~\cite{FSCIL-3D}}} & train & 26287 & 64 & 69 & 74 & 79 & 84 & 89 &94 & 99 & 104 & 109 & - \\
        \multicolumn{1}{l}{} & test & 6604 & 6793 & 6959 & 7205 & 7333 & 7492 & 7646 & 7816 & 7957 & 8117 & 8336 & - \\
        \midrule
        \multicolumn{1}{l}{\multirow{2}{*}{S2S}} & train & 41943 & 75 & 79 & 83 & 87 & 91 & 95 & - & - & - & - & - \\
        \multicolumn{1}{l}{} & test & 10517 & 10663 & 10889 & 11195 & 11435 & 11595 & 11835 & - & - & - & - & - \\
        \midrule
        \multicolumn{1}{l}{\multirow{2}{*}{S2R}} & train & 41943 & 75 & 79 & 83 & 87 & 91 & 95 & 99 & 103 & 107 & 111 & 100 \\
        \multicolumn{1}{l}{} & test & 10517 & 10867 & 11020 & 11395 & 11705 & 11897 & 12226 & 12505 & 12718 & 13022 & 13313 & 13404 \\
        \bottomrule
  \end{tabular}
  }
  \vspace{-0.2em}
  \label{tab-samples}
\end{table*}

\subsection{Implementation Details}
We select CLIP's ViT-B/32~\cite{CLIP} as our pre-trained model, replacing its visual encoder with CLIP2Point's pre-trained depth encoder~\cite{CLIP2Point} and adopting CLIP2Point's proposed rendering approach. \textcolor{red}{The number of views $N$ is set to 6.} For the 3D encoder, we utilize DGCNN~\cite{DGCNN} pretrained on ShapeNet, following the methodology of OcCo~\cite{OcCo}.
Both image and text embedding dimensions $C$ are 512. We employ SVD~\cite{svd} as our dimensionality reduction technique, extracting 242 principal components based on the base task (\textit{i.e.} $M$ = 242), which retains 95\% energy of the principal components. The temperature parameter of infoNCE and $\mathcal{L}_{align}$ are set to 0.7 and 1.0 respectively. The balance parameter $\alpha$ and $\beta$ in $\mathcal{L}_{align}$ are set to 0.25.
For training, we utilize the ADAM weight decay optimizer~\cite{adamw}, with the learning rate set to $1 \times 10^{-3}$ and the weight decay to $1 \times 10^{-4}$. Training in the base task and incremental tasks lasts for 10 and 20 epochs, respectively. In incremental tasks, we randomly select 5 samples/classes for training and allow 1 exemplar/class from previous tasks to be utilized as memory. \textcolor{red}{The number of samples $n_b$ in the training and testing sets for the S2S and S2R tasks is shown in Table ~\ref{tab-samples}.}
The batch size is set to 32. 
\textcolor{red}{All of our experiments are conducted on 1 A6000 GPU using PyTorch. }

\subsection{Computational Cost}
\textcolor{red}{An iteration (it) processes one batch, which consists of 32 samples. During the training of base classes, the GPU memory consumption is 7814 MiB, with an average inference speed of 1.34 it/s. During incremental training, the GPU memory consumption increases to 12838 MiB, with an average inference speed of 3.41 s/it. During test inference, the GPU memory consumption remains at 7776 MiB, with an average inference speed of 2.65 it/s.}


\subsection{Experimental Results}
\noindent\textbf{Compared methods.}
To make a comprehensive comparison, we choose the following models: 1) $Joint$: Models are trained with samples from all currently visible classes, which can represent their upper bound in FSCIL tasks. 2) One 3D FSCIL SOTA approach developed by \citet{FSCIL-3D}. 3) Two 3D few-shot SOTA approaches PointCLIP~\cite{zhang2022pointclip} and ULIP~\cite{ULIP} which also utilize CLIP as backbone. We compare our model with these two models to validate the effectiveness of our model's design in addressing continuous-learning-related issues. We further introduce PointCLIP++ (PointCLIP backbone + RFE + SNC) to validate the generalization of our proposed modules. 4) Two 2D SOTA approaches FACT~\cite{FACT} and BiDist~\cite{BiDist}. FACT uses manifold-mixup~\cite{maniflod-mixup} and virtual prototypes to ensure forward compatibility. BiDist uses knowledge distillation to retain knowledge. We use official codes to reproduce these methods, keeping all the original parameters unchanged. The only difference is that we replace the CNN-based network with CLIP2Point's depth encoder to allow 3D application. 

Notice that joint training suffers from the long-tail issue on new classes but can be properly solved on the FSCIL framework. Therefore, joint training results are not exactly the upper limit of $\rm NCAcc$ and $\rm F$. Therefore we exclude $\rm NCAcc, F$ metrics of $joint$ in the experiment table.

\begin{table*}[!t]
  \caption{Quantitative results on \citet{FSCIL-3D}'s ShapeNet2CO3D benchmark. \textbf{Bold} denotes the best performance, $joint$ serves solely as an upper reference limit in \citet{FSCIL-3D}'s work and is not involved in the comparison.}
  \centering
  \resizebox{0.99\linewidth}{!}{
  \begin{tabular}{lcccccccccccc}     
        \toprule
        \multicolumn{1}{l}{\multirow{2}{*}{Method}} & \multicolumn{11}{c}{Acc. in each session $\uparrow$ } & \multicolumn{1}{c}{\multirow{2}{*}{${\rm \Delta}\downarrow$}}  \\
        \cmidrule{2-12}
        \multicolumn{1}{l}{} & 0 & 1 & 2 & 3 & 4 & 5 & 6 & 7 & 8 & 9 & 10 & \multicolumn{1}{c}{} \\
        \midrule
        \multicolumn{1}{l}{$Joint$ Chowdhury et al.'s~\cite{FSCIL-3D}} & 81.0 & 79.5 & 78.3 & 75.2 & 75.1 & 74.8 & 72.3 & 71.3 & 70.0 & 68.8 & 67.3 & 16.9 \\
        \midrule
        \multicolumn{1}{l}{FACT~\cite{FACT}} & 81.4 & 76.0 & 70.3 & 68.1 & 65.8 & 63.5 & 63.0 & 60.1 & 58.2 & 57.5 & 55.9 & 31.3 \\
        \multicolumn{1}{l}{Chowdhury et al.’s ~\cite{FSCIL-3D}} & 82.6 & 77.9 & 73.9 & 72.7 & 67.7 & 66.2 & 65.4 & 63.4 & 60.6 & 58.1 & 57.1 & 30.9 \\
        \midrule
        \multicolumn{1}{l}{\textcolor{red}{baseline (ours)}} & \textcolor{red}{85.4} & \textcolor{red}{81.3} & \textcolor{red}{80.1} & \textcolor{red}{77.9} & \textcolor{red}{75.1} & \textcolor{red}{73.8} & \textcolor{red}{72.3} & \textcolor{red}{70.6} & \textcolor{red}{69.3} & \textcolor{red}{67.1} & \textcolor{red}{65.6} & \textcolor{red}{23.2} \\   
        \multicolumn{1}{l}{FILP-3D (ours)} & \textbf{87.3} & \textbf{81.5} & \textbf{82.3} & \textbf{80.1} & \textbf{78.5} & \textbf{76.4} & \textbf{76.8} & \textbf{74.2} & \textbf{71.6} & \textbf{71.7} & \textbf{70.1} & \textbf{19.7} \\
        \bottomrule
  \end{tabular}
  }
  \label{tab-fscil}
\end{table*}
\noindent\textbf{Experiments on \citet{FSCIL-3D}'s benchmark.}
We report the performance on \citet{FSCIL-3D}'s ShapeNet2CO3D benchmark in Table~\ref{tab-fscil}. \citet{FSCIL-3D}'s work does not provide any detailed data partitions other than ShapeNet2CO3D. Consequently, we are unable to follow them and conduct further experiments and comparisons, such as incremental within a single dataset. As shown in Table~\ref{tab-fscil}, FILP-3D outperforms the other two methods. After the final task, our accuracy only drops to 70.1\%, which surpasses the upper bound of \citet{FSCIL-3D}'s method. This demonstrates the vast potential of CLIP in the 3D FSCIL task. However, we note that the results in this benchmark cannot distinctly indicate whether the performance degradation of ~\citet{FSCIL-3D} is caused by the new class or the old classes. We cannot conclude more observations based on it.

\begin{table*}[!t]
  \caption{Quantitative results on the S2S task. For each set of results, the micro/macro average are presented at the top/bottom respectively. \textbf{Bold} denotes the best performance, $joint$ serves solely as an upper reference limit in our model and is not involved in the comparison.}
  \centering
  \resizebox{0.99\linewidth}{!}{
  \begin{tabular}{lccccccccccc}
        \toprule
        \multicolumn{1}{l}{\multirow{2}{*}{Method}} & \multicolumn{1}{l}{\multirow{2}{*}{Pub. Year}} & \multicolumn{7}{c}{Acc. in each session $\uparrow$} & \multicolumn{3}{c}{Evaluation metrics}  \\
        \cmidrule{3-12}
        \multicolumn{1}{l}{} & \multicolumn{1}{l}{} & 0 & 1 & 2 & 3 & 4 & 5 & 6 & ${\rm NCAcc}\uparrow$ & ${\rm \Delta}\downarrow$ & ${\rm F}\uparrow$ \\
        \midrule
        \multicolumn{1}{l}{\multirow{2}{*}{FACT~\cite{FACT}}} & \multicolumn{1}{c}{\multirow{2}{*}{CVPR'22}} & 82.6 & 77.0 & 72.4 & 69.8 & 68.4 & 67.7 & 67.3 & 41.7 & 18.5 & 51.5 \\
        \multicolumn{1}{l}{} & \multicolumn{1}{l}{} & 48.0 & 44.7 & 42.0 & 39.9 & 38.2 & 37.1 & 36.5 & 34.0 & 23.9 & 35.2 \\
        \midrule
        \multicolumn{1}{l}{\multirow{2}{*}{BiDist~\cite{BiDist}}} & \multicolumn{1}{c}{\multirow{2}{*}{CVPR'23}} & 89.6 & 87.7 & 86.2 & 84.7 & 83.8 & \textbf{83.6} & \textbf{82.3} & 35.0 & \textbf{8.1} & 49.1 \\
        \multicolumn{1}{l}{} & \multicolumn{1}{l}{} & \textbf{82.5} & \textbf{80.3} & \textbf{77.3} & \textbf{75.2} & \textbf{72.5} & \textbf{71.2} & 69.3 & 38.0 & 16.0 & 49.1 \\
        \midrule
        \multicolumn{1}{l}{\multirow{2}{*}{Chowdhury et al.~\cite{FSCIL-3D}}} & \multicolumn{1}{c}{\multirow{2}{*}{ECCV'22}} & 86.9 & 84.6 & 82.8 & 78.3 & 78.5 & 71.5 & 68.6 & 50.8 & 21.1 & 58.4 \\
        \multicolumn{1}{l}{} & \multicolumn{1}{l}{} & 73.0 & 62.8 & 65.2 & 62.8 & 60.9 & 57.9 & 55.1 & 45.5 & 24.5 & 49.8 \\
        \midrule
        \multicolumn{1}{l}{\multirow{2}{*}{ULIP~\cite{ULIP}}} & \multicolumn{1}{c}{\multirow{2}{*}{CVPR'23}} & 86.3 & 83.3 & 80.3 & 75.8 & 72.8 & 62.9 & 65.1 & 78.3 & 24.6 & 71.1 \\
        \multicolumn{1}{l}{} & \multicolumn{1}{l}{} & 85.4 & 80.5 & 76.3 & 71.4 & 69.3 & 67.6 & 64.9 & \textbf{77.2} & 24.0 & 70.5\\
        \midrule
        \multicolumn{1}{l}{\multirow{2}{*}{PointCLIP~\cite{zhang2022pointclip}}} & \multicolumn{1}{c}{\multirow{2}{*}{CVPR'22}} & 88.3 & 85.8 & 81.0 & 73.9 & 72.0 & 71.9 & 67.7 & 54.3 & 23.3 & 60.3\\
        \multicolumn{1}{l}{} & \multicolumn{1}{l}{} & 75.8 & 71.8 & 67.5 & 63.6 & 62.1 & 59.9 & 56.9 & 54.9 & 24.9 & 55.9\\
        \midrule
        \multicolumn{1}{l}{\multirow{2}{*}{PointCLIP++}} & \multicolumn{1}{c}{\multirow{2}{*}{-}} & 88.8 & 86.3 & 82.3 & 80.0 & 77.5 & 74.0 & 70.3 & 60.7 & 20.8 & 65.1 \\
        \multicolumn{1}{l}{} & \multicolumn{1}{l}{} & 76.2 & 71.7 & 68.9 & 67.5 & 64.2 & 61.1 & 59.8 & 61.6 & 21.5 & 60.7\\
        \midrule
        \midrule
        \multicolumn{1}{l}{\multirow{2}{*}{baseline (ours)}} & \multicolumn{1}{c}{\multirow{2}{*}{-}} & 90.4 & 88.0 & 85.6 & \textbf{85.0} & \textbf{84.1} & 78.4 & 80.1 & 68.2 & 11.4 & 73.7 \\
        \multicolumn{1}{l}{} & \multicolumn{1}{l}{} & 78.9 & 76.1 & 72.8 & 71.1 & 70.8 & 70.2 & 69.7 & 68.8 & 11.7 & 69.2 \\
        \midrule
        \multicolumn{1}{l}{\multirow{2}{*}{FILP-3D (ours)}} & \multicolumn{1}{c}{\multirow{2}{*}{-}} & \textbf{90.6} & \textbf{89.0} & \textbf{86.7} & 84.2 & 83.2 & 81.8 & 82.2 & \textbf{79.3} & 9.3 & \textbf{80.7} \\
        \multicolumn{1}{l}{} & \multicolumn{1}{l}{} & 80.0 & 76.8 & 74.9 & 72.8 & 71.2 & 70.9 & \textbf{70.7} & 77.0 & \textbf{11.6} & \textbf{73.7} \\
        \midrule
        \midrule
        \multicolumn{1}{l}{\multirow{2}{*}{$Joint$ FILP-3D}} & \multicolumn{1}{c}{\multirow{2}{*}{-}} & 90.6 & 89.1 & 88.5 & 87.8 & 87.4 & 87.6 & 86.8 & - & 4.2 & - \\
        \multicolumn{1}{l}{} & \multicolumn{1}{l}{} & 80.0 & 78.5 & 79.0 & 77.2 & 77.5 & 77.7 & 77.3 & - & 3.4 & - \\
        \bottomrule
  \end{tabular}
  }
  \vspace{-0.2em}
  \label{tab-shapenet2modelnet}
\end{table*}

\noindent\textbf{Experiments on the S2S task of FSCIL3D-XL.}
\textcolor{red}{Table~\ref{tab-shapenet2modelnet} reports the results of the synthetic data to the synthetic data task in our benchmark, whilch mainly reflect models' capability to mitigate issues like overfitting or catastrophic forgetting. }

\textcolor{red}{FACT~\cite{FACT}, \citet{FSCIL-3D}'s work, and ULIP lack the shape-relative prior knowledge and therefore perform worse than methods that use CLIP as the backbone, such as PointCLIP~\cite{zhang2022pointclip} and baseline. This demonstrates that V-L PTMs are effective in addressing the dual challenges of forgetting and overfitting, not only in 2D FSCIL tasks but also in 3D FSCIL tasks. }

BiDist~\cite{BiDist} additionally employs a weighted sum of distillation loss for optimization, along with a relatively higher loss weight assigned to it. Therefore, BiDist tends to preserve existing knowledge and sacrifices the learning capacity for new classes, leading to the highest $\rm Acc$ and lowest $\rm NCAcc$. Contrary to it, FILP-3D has a trade-off between the base classes and the novel classes. The results also show that our metrics are reasonable for the evaluation of 3D FSCIL. 

\textcolor{red}{The PointCLIP~\cite{zhang2022pointclip} architecture is similar to the baseline, hence they exhibit comparable performance. PointCLIP++ and FILP-3D show significant performance improvements over PointCLIP and the baseline respectively, particularly for new classes ($\rm{NCAcc}$). This phenomenon highlights the importance of the feature space alignment issue addressed by RFE, which plays a crucial role in enabling V-L PTMs to fully leverage their potential in 3D FSCIL tasks.}

\begin{table*}[!t]
  \caption{Quantitative results on the S2R task. For each set of results, the micro/macro average are presented at the top/bottom respectively. \textbf{Bold} denotes the best performance, $joint$ serves solely as an upper reference limit in our model and is not involved in the comparison.}
  \centering
  \resizebox{0.99\linewidth}{!}{
  \begin{tabular}{lccccccccccccccc}
        \toprule
        \multicolumn{1}{l}{\multirow{2}{*}{Method}} & \multicolumn{12}{c}{Acc. in each session $\uparrow$} & \multicolumn{3}{c}{Evaluation metrics}  \\
        \cmidrule{2-16}
        \multicolumn{1}{l}{} & 0\hspace{-1mm} & 1\hspace{-1mm} & 2\hspace{-1mm} & 3\hspace{-1mm} & 4\hspace{-1mm} & 5\hspace{-1mm} & 6\hspace{-1mm} & 7\hspace{-1mm} & 8\hspace{-1mm} & 9\hspace{-1mm} & 10\hspace{-1mm} & 11\hspace{-1mm} & ${\rm NCAcc}\uparrow$ & ${\rm \Delta}\downarrow$ & ${\rm F}\uparrow$ \\
        \midrule
        \multicolumn{1}{l}{\multirow{2}{*}{FACT~\cite{FACT}}} & 82.4\hspace{-1mm} & 77.2\hspace{-1mm} & 74.5\hspace{-1mm} & 73.1\hspace{-1mm} & 71.3\hspace{-1mm} & 70.4\hspace{-1mm} & 67.2\hspace{-1mm} & 65.2\hspace{-1mm} & 63.8\hspace{-1mm} & 61.8\hspace{-1mm} & 59.9\hspace{-1mm} & 59.8\hspace{-1mm} & 26.2 & 27.4 & 36.4 \\
        \multicolumn{1}{l}{} & 48.6\hspace{-1mm} & 41.4\hspace{-1mm} & 39.7\hspace{-1mm} & 36.8\hspace{-1mm} & 35.5\hspace{-1mm} & 33.6\hspace{-1mm} & 31.2\hspace{-1mm} & 29.5\hspace{-1mm} & 28.4\hspace{-1mm} & 27.2\hspace{-1mm} & 25.8\hspace{-1mm} & 25.9\hspace{-1mm} & 30.8 & 46.7 & 28.1\\  
        \midrule
        \multicolumn{1}{l}{\multirow{2}{*}{BiDist~\cite{BiDist}}} & 89.4\hspace{-1mm} & 54.0\hspace{-1mm} & 54.7\hspace{-1mm} & 56.4\hspace{-1mm} & 57.0\hspace{-1mm} & 55.9\hspace{-1mm} & 56.3\hspace{-1mm} & 52.9\hspace{-1mm} & 52.3\hspace{-1mm} & 51.7\hspace{-1mm} & 50.8\hspace{-1mm} & 50.1\hspace{-1mm} & 47.2 & 43.9 & 48.6 \\
        \multicolumn{1}{l}{} & \textbf{81.8}\hspace{-1mm} & 52.8\hspace{-1mm} & 50.1\hspace{-1mm} & 46.2\hspace{-1mm} & 48.3\hspace{-1mm} & 46.1\hspace{-1mm} & 44.7\hspace{-1mm} & 41.8\hspace{-1mm} & 41.8\hspace{-1mm} & 39.8\hspace{-1mm} & 40.0\hspace{-1mm} & 39.7\hspace{-1mm} & 41.9 & 51.5 & 40.8 \\   
        \midrule
        \multicolumn{1}{l}{\multirow{2}{*}{Chowdhury et al.~\cite{FSCIL-3D}}} & 85.2\hspace{-1mm} & 78.6\hspace{-1mm} & 71.0\hspace{-1mm} & 72.0\hspace{-1mm} & 75.2\hspace{-1mm} & 68.8\hspace{-1mm} & 56.1\hspace{-1mm} & 58.5\hspace{-1mm} & 62.9\hspace{-1mm} & 59.1\hspace{-1mm} & 52.2\hspace{-1mm} & 59.4\hspace{-1mm} & 35.3 & 30.3 & 44.3 \\
        \multicolumn{1}{l}{} & 68.2\hspace{-1mm} & 56.2\hspace{-1mm} & 50.5\hspace{-1mm} & 48.4\hspace{-1mm} & 53.5\hspace{-1mm} & 46.7\hspace{-1mm} & 39.9\hspace{-1mm} & 37.6\hspace{-1mm} & 36.9\hspace{-1mm} & 33.1\hspace{-1mm} & 34.3\hspace{-1mm} & 44.1\hspace{-1mm} & 36.5 & 35.3 & 40.0\\
        \midrule
        \multicolumn{1}{l}{\multirow{2}{*}{ULIP~\cite{ULIP}}} & 86.3\hspace{-1mm} & 85.6\hspace{-1mm} & 81.7\hspace{-1mm} & 74.0\hspace{-1mm} & 71.7\hspace{-1mm} & 68.1\hspace{-1mm} & 67.6\hspace{-1mm} & 64.5\hspace{-1mm} & 59.5\hspace{-1mm} & 58.4\hspace{-1mm} & 55.2\hspace{-1mm} & 57.5\hspace{-1mm} & 56.4 & 33.4 & 56.9\\
        \multicolumn{1}{l}{} & \textbf{85.4}\hspace{-1mm} & \textbf{82.0}\hspace{-1mm} & \textbf{78.2}\hspace{-1mm} & 70.4\hspace{-1mm} & 66.9\hspace{-1mm} & 63.3\hspace{-1mm} & 61.9\hspace{-1mm} & 58.1\hspace{-1mm} & 54.8\hspace{-1mm} & 52.1\hspace{-1mm} & 49.2\hspace{-1mm} & 50.3\hspace{-1mm} & 57.0 & 41.1 & 53.4\\
        \midrule
        \multicolumn{1}{l}{\multirow{2}{*}{PointCLIP~\cite{zhang2022pointclip}}} & 87.5\hspace{-1mm} & 83.2\hspace{-1mm} & 79.5\hspace{-1mm} & 74.9\hspace{-1mm} & 75.6\hspace{-1mm} & 74.2\hspace{-1mm} & 71.1\hspace{-1mm} & 71.7\hspace{-1mm} & 67.9\hspace{-1mm} & 66.5\hspace{-1mm} & 64.3\hspace{-1mm} & 66.1\hspace{-1mm} & 54.1 & 24.5 & 59.5\\
        \multicolumn{1}{l}{} & 74.2\hspace{-1mm} & 72.1\hspace{-1mm} & 68.8\hspace{-1mm} & 62.3\hspace{-1mm} & 60.1\hspace{-1mm} & 57.9\hspace{-1mm} & 54.5\hspace{-1mm} & 55.1\hspace{-1mm} & 50.9\hspace{-1mm} & 50.8\hspace{-1mm} & 47.5\hspace{-1mm} & 48.7\hspace{-1mm} & 53.2 & 34.4 & 50.9\\
        \midrule
        \multicolumn{1}{l}{\multirow{2}{*}{PointCLIP++}} & 88.2\hspace{-1mm} & 85.6\hspace{-1mm} & 84.2\hspace{-1mm} & 81.9\hspace{-1mm} & 80.2\hspace{-1mm} & 78.4\hspace{-1mm} & 75.3\hspace{-1mm} & 72.6\hspace{-1mm} & 71.5\hspace{-1mm} & 69.3\hspace{-1mm} & 67.3\hspace{-1mm} & 68.1\hspace{-1mm} & 59.8 & 22.7 & 63.7\\
        \multicolumn{1}{l}{} & 75.1\hspace{-1mm} & 72.6\hspace{-1mm} & 70.4\hspace{-1mm} & 65.4\hspace{-1mm} & 63.3\hspace{-1mm} & 61.0\hspace{-1mm} & 57.3\hspace{-1mm} & 53.9\hspace{-1mm} & 53.6\hspace{-1mm} & 51.4\hspace{-1mm} & 49.6\hspace{-1mm} & 50.3\hspace{-1mm} & 59.7 & 33.0 & 54.6\\
        \midrule
        \midrule
        \multicolumn{1}{l}{\multirow{2}{*}{baseline (ours)}} & 89.2\hspace{-1mm} & 86.7\hspace{-1mm} & 83.5\hspace{-1mm} & 81.7\hspace{-1mm} & 79.4\hspace{-1mm} & 79.6\hspace{-1mm} & 78.6\hspace{-1mm} & 70.4\hspace{-1mm} & 72.1\hspace{-1mm} & 71.7\hspace{-1mm} & 70.1\hspace{-1mm} & 71.2\hspace{-1mm} & 49.6 & 20.2 &  58.5\\
        \multicolumn{1}{l}{} & 78.9\hspace{-1mm} & 77.3\hspace{-1mm} & 75.9\hspace{-1mm} & \textbf{73.7}\hspace{-1mm} & \textbf{70.1}\hspace{-1mm} & 66.5\hspace{-1mm} & 64.4\hspace{-1mm} & 60.9\hspace{-1mm} & 59.4\hspace{-1mm} & 58.5\hspace{-1mm} & 54.5\hspace{-1mm} & 56.3\hspace{-1mm} & 49.0 & 28.6 & 52.4\\
        \midrule
        \multicolumn{1}{l}{\multirow{2}{*}{FILP-3D (ours)}} & \textbf{90.2}\hspace{-1mm} & \textbf{87.1}\hspace{-1mm} & \textbf{85.9}\hspace{-1mm} & \textbf{84.4}\hspace{-1mm} & \textbf{83.4}\hspace{-1mm} & \textbf{82.3}\hspace{-1mm} & \textbf{80.4}\hspace{-1mm} & \textbf{79.3}\hspace{-1mm} & \textbf{78.1}\hspace{-1mm} & \textbf{76.7}\hspace{-1mm} & \textbf{74.6}\hspace{-1mm} & \textbf{74.4}\hspace{-1mm} & \textbf{60.4} & \textbf{17.5} & \textbf{66.7} \\
        \multicolumn{1}{l}{} & 79.5\hspace{-1mm} & 75.3\hspace{-1mm} & 74.8\hspace{-1mm} & 71.6\hspace{-1mm} & 69.6\hspace{-1mm} & \textbf{67.9}\hspace{-1mm} & \textbf{65.3}\hspace{-1mm} & \textbf{62.7}\hspace{-1mm} & \textbf{61.7}\hspace{-1mm} & \textbf{60.0}\hspace{-1mm} & \textbf{58.2}\hspace{-1mm} & \textbf{57.7}\hspace{-1mm} & \textbf{61.8} & \textbf{27.4} & \textbf{59.7}\\
        \midrule
        \midrule
        \multicolumn{1}{l}{\multirow{2}{*}{$Joint$ FILP-3D}} & 90.0\hspace{-1mm} & 89.2\hspace{-1mm} & 89.0\hspace{-1mm} & 88.4\hspace{-1mm} & 88.1\hspace{-1mm} & 87.7\hspace{-1mm} & 87.2\hspace{-1mm} & 86.9\hspace{-1mm} & 85.9\hspace{-1mm} & 84.6\hspace{-1mm} & 83.1\hspace{-1mm} & 83.1\hspace{-1mm} & - & 7.7 & - \\
        \multicolumn{1}{l}{} & 79.4\hspace{-1mm} & 77.2\hspace{-1mm} & 76.3\hspace{-1mm} & 75.9\hspace{-1mm} & 75.5\hspace{-1mm} & 73.5\hspace{-1mm} & 73.3\hspace{-1mm} & 73.4\hspace{-1mm} & 70.2\hspace{-1mm} & 69.3\hspace{-1mm} & 68.2\hspace{-1mm} & 68.2\hspace{-1mm} & - & 14.1 & - \\  
        \bottomrule
  \end{tabular}
  }
  \label{tab-shapenet2co3d}
  \vspace{-1.0em}
\end{table*}

\noindent\textbf{Experiments on the S2R task of FSCIL3D-XL.}
Table~\ref{tab-shapenet2co3d} reports the results of the synthetic data to the real-scanned data task in our benchmark. Considering the large domain gap between synthetic data and real-scanned data, it is even harder to acquire knowledge from new classes based on previous knowledge of synthetic shapes. 

By observing the experiment results presented in Table~\ref{tab-shapenet2co3d}, we notice a catastrophic decline in FACT~\cite{FACT}'s performance in terms of $\rm MAcc$ when learning real-scanned data. This may be because The feature space trained by the manifold-mixup~\cite{maniflod-mixup} method is relatively fragile in the aspect of structure, thus it is easy to overfit the noise information during the incremental process, especially for real-scanned data. Solely relying on metrics proposed by Chowdhury et al., we are unable to identify the aforementioned issues and then perform the analysis.

We also notice that, unlike the experiment on the S2S task, BiDist~\cite{BiDist} witnesses a catastrophic performance decline from task 1 to task 2, which further demonstrates that BiDist is unable to balance the retaining of old knowledge and the acquisition of new knowledge in scenarios with a large domain gap. In contrast, our baseline has already outperformed the other three methods, which proves that prior knowledge from pre-training models is effective for few-shot incremental tasks. 

Comparing ULIP with our model, it's evident that ULIP exhibits outstanding learning capabilities in novel classes. However, the abundance of trainable parameters and severely inadequate data make it extremely challenging to avoid overfitting during the FSCIL process, ultimately resulting in a high dropping rate $\rm \Delta$ and limited performance. Consequently, we conclude that a frozen 2D PTM as the backbone is more suitable for 3D FSCIL. 

The PointCLIP also exhibits comparable performance to the baseline because of similar architecture. Experimental results demonstrate that PointCLIP++ outperforms PointCLIP significantly across various metrics, especially in $\rm NCAcc$. This validates the generality and effectiveness of our proposed modules.

For our FILP-3D, the 2D PTM and simple adapters provide FILP-3D with good generality and adaptability. On the other hand, RFE and SNC focus on addressing specific issues related to applying multi-view 3D models to FSCIL tasks, thereby achieving SOTA performance.

\noindent\textbf{Effect of our proposed metrics.}
By observing the experiment results presented in Table~\ref{tab-shapenet2modelnet}, a notable deterioration in Chowdhury et al.'s~\cite{FSCIL-3D} performance is observed concerning $\rm MAcc$ during incremental learning, accompanied by subpar results in the $\rm NCAcc$ metric. This indicates their ineffectiveness in acquiring new knowledge efficiently. Solely relying on micro Acc. and dropping rate $\Delta$, we are unable to identify the aforementioned issues. Similarly, without $\rm NCAcc$, Bidist's inability to effectively learn new classes also can not be revealed. Similarly, without the $\rm NCAcc$ metric in Table~\ref{tab-shapenet2co3d}, Chowdhury et al.'s terrible performance in learning real scanned new classes also can not be noticed. 

Consequently, our proposed metrics, $\rm MAcc$ and $\rm NCAcc$, hold significant importance in comprehensively evaluating and analyzing the 3D FSCIL model. Moreover, ${\rm F_{FSCIL}}$, serving as a metric that balances learning new classes and not forgetting old ones, can objectively evaluate the performance of a continual learning model as much as possible.

\begin{table}[!t]
  \caption{Compare RFE with other methods.}
  \centering
  \resizebox{0.5\linewidth}{!}{
  \begin{tabular}{ccccccc}
        \toprule
        Method & 0 & 11 & ${\rm NCAcc}\uparrow$ & ${\rm \Delta}\downarrow$ & ${\rm F}\uparrow$ \\ 
        \midrule
        \multicolumn{1}{c}{\multirow{2}{*}{APE}} & \textbf{90.3} & 72.8 & 53.8 & 19.4 & 61.9\\
        \multicolumn{1}{l}{} & 79.3 & 56.5 & 55.7 & 28.8 & 56.1\\
        \midrule
        \multicolumn{1}{c}{\multirow{2}{*}{DomainDrop}} & \textbf{90.3} & 72.7 & 54.4 & 19.5 & 62.2\\
        \multicolumn{1}{l}{} & \textbf{79.9} & 55.9 & 56.1 & 30.0 & 56.0\\
        \midrule
        \multicolumn{1}{c}{\multirow{2}{*}{FILP-3D}} & 90.2 & \textbf{74.4} & \textbf{60.4} & \textbf{17.5} & \textbf{66.7} \\
        \multicolumn{1}{c}{} & 79.5 & \textbf{57.7} & \textbf{61.8} & \textbf{27.4} & \textbf{59.6} \\
        \bottomrule
  \end{tabular}
  }
  \label{compare}
\end{table}

\noindent\textbf{Comparsion between RFE with other channel mask methods.}
We conduct comparison experiments between RFE with APE~\cite{zhu2023not} and DomainDrop~\cite{guo2023domaindrop} in the S2R task to reveal the superiority of RFE. APE utilizes textual features of classes to filter out redundant channels. However, the number of classes in the base task (55) is limited compared to APE (1000), making it prone to mistakenly masking semantic features. The random drop of domain-sensitive channels proposed by DomainDrop with a 33\% drop rate is too aggressive and unstable for the few-shot and incremental setting, leading to overfitting and forgetting. These methods are primarily tailored for domain adaptation tasks and consequently exhibit limited performance due to their failure to address the challenges inherent in FSCIL. Contrary to them, our proposed RFE fully leverages abundant data from base classes to preprocess more general principal components, combined with the RCS to avoid DR compromising semantic integrity. According to Table~\ref{compare}, our RFE outperforms APE and DomainDrop in terms of almost all metrics.

\begin{minipage}[h]{1.0\textwidth}
\begin{minipage}[h]{0.52\textwidth}
    \makeatletter\def\@captype{table}
    \captionsetup{font=scriptsize}
    \centering
    \caption{Ablation studies on the S2S task of FSCIL3D-XL. \textcolor{red}{The ablation experiments for RFE and RCS were conducted in this set of experiments.}}
    \label{tab: ablation}
    \vspace{-0.4cm}
    \resizebox{1.0\linewidth}{!}{
    \begin{tabular}{ccccccc}
    \toprule
        \multicolumn{1}{l}{DR} & \multicolumn{1}{l}{RCS} & 0 & 6 & ${\rm NCAcc}\uparrow$ & ${\rm \Delta}\downarrow$ & ${\rm F}\uparrow$ \\
        \midrule
        \multicolumn{1}{l}{\multirow{2}{*}{\XSolidBrush}} & \multicolumn{1}{c}{\multirow{2}{*}{\XSolidBrush}} & 90.4 & 80.1 & 68.2 & 11.4 & 73.7 \\
        \multicolumn{1}{l}{} & \multicolumn{1}{c}{} & 78.9 & 69.7 & 68.8 & 11.7 & 69.2 \\
        \midrule
        \multicolumn{1}{l}{\multirow{2}{*}{\Checkmark}} & \multicolumn{1}{c}{\multirow{2}{*}{\XSolidBrush}} & 90.5 & 81.2 & 70.9 & 10.2 & 75.7 \\
        \multicolumn{1}{l}{} & \multicolumn{1}{c}{} & 79.7 & 69.7 & 70.9 & 12.5 & 70.3 \\
        \midrule
        \multicolumn{1}{l}{\multirow{2}{*}{\Checkmark}} & \multicolumn{1}{c}{\multirow{2}{*}{\Checkmark}} & 90.6 & 82.2 & 79.3 & 9.3 & 80.7 \\
        \multicolumn{1}{l}{} & \multicolumn{1}{c}{} & 80.0 & 70.7 & 77.0 & 11.6 & 73.7 \\
        \bottomrule
    \end{tabular}%
    }
    \end{minipage}
    \begin{minipage}[h]{0.45\textwidth}
    \makeatletter\def\@captype{table}
    \captionsetup{font=scriptsize}
    \centering
    \caption{Ablation studies on the S2R task of FSCIL3D-XL. \textcolor{red}{The ablation experiments for the RFE module, the SNC module, and contrastive learning. were conducted in this set of experiments.}}
    \vspace{-0.4cm}
    \resizebox{0.9\linewidth}{!}{
    \label{tab: ablation2}
    \begin{tabular}{cccccccc}
    \toprule
        \multicolumn{1}{c}{RFE} & \multicolumn{1}{c}{SNC} & \multicolumn{1}{c}{CL} & 0\hspace{-1mm} & 11\hspace{-1mm} & ${\rm NCAcc}\uparrow$\hspace{-1mm} & ${\rm \Delta}\downarrow$\hspace{-1mm} & ${\rm F}\uparrow$\hspace{-1mm} \\
        \midrule
        \multicolumn{1}{c}{\multirow{2}{*}{\XSolidBrush}} & \multicolumn{1}{c}{\multirow{2}{*}{\XSolidBrush}} & \multicolumn{1}{c}{\multirow{2}{*}{\XSolidBrush}} & 86.7\hspace{-1mm} & 71.2\hspace{-1mm} & 49.6\hspace{-1mm} & 17.9\hspace{-1mm} & 58.5\hspace{-1mm} \\
        \multicolumn{1}{c}{} & \multicolumn{1}{c}{} & \multicolumn{1}{c}{} & 78.9\hspace{-1mm} & 56.3\hspace{-1mm} & 49.0\hspace{-1mm} & 28.6\hspace{-1mm} & 52.4\hspace{-1mm} \\
        \midrule
        \multicolumn{1}{c}{\multirow{2}{*}{\Checkmark}} & \multicolumn{1}{c}{\multirow{2}{*}{\XSolidBrush}} & \multicolumn{1}{c}{\multirow{2}{*}{\XSolidBrush}} & 90.4\hspace{-1mm} & 75.6\hspace{-1mm} & 50.8\hspace{-1mm} & 16.3\hspace{-1mm} & 60.8\hspace{-1mm} \\
        \multicolumn{1}{c}{} & \multicolumn{1}{c}{} & \multicolumn{1}{c}{}  & 80.0\hspace{-1mm} & 58.3\hspace{-1mm} & 51.0\hspace{-1mm} & 27.1\hspace{-1mm} & 54.4\hspace{-1mm}\hspace{-1mm} \\
        \midrule
        \multicolumn{1}{c}{\multirow{2}{*}{\XSolidBrush}} & \multicolumn{1}{c}{\multirow{2}{*}{\Checkmark}} & \multicolumn{1}{c}{\multirow{2}{*}{\XSolidBrush}} & 90.1\hspace{-1mm} & 73.6\hspace{-1mm} & 54.5\hspace{-1mm} & 18.3\hspace{-1mm} & 62.6\hspace{-1mm} \\
        \multicolumn{1}{c}{} & \multicolumn{1}{c}{} & \multicolumn{1}{c}{}  & 79.3\hspace{-1mm} & 56.7\hspace{-1mm} & 56.1\hspace{-1mm} & 28.5\hspace{-1mm} & 56.4\hspace{-1mm} \\
        \midrule
        \multicolumn{1}{c}{\multirow{2}{*}{\Checkmark}} & \multicolumn{1}{c}{\multirow{2}{*}{\Checkmark}} & \multicolumn{1}{c}{\multirow{2}{*}{\XSolidBrush}} & 89.8\hspace{-1mm} & 74.0\hspace{-1mm} & 57.8\hspace{-1mm} & 17.6\hspace{-1mm} & 64.9\hspace{-1mm} \\
        \multicolumn{1}{c}{} & \multicolumn{1}{c}{} & \multicolumn{1}{c}{}  & 78.7\hspace{-1mm} & 56.7\hspace{-1mm} & 58.0\hspace{-1mm} & 28.0\hspace{-1mm} & 57.3\hspace{-1mm} \\
        \midrule
        \multicolumn{1}{c}{\multirow{2}{*}{\Checkmark}} & \multicolumn{1}{c}{\multirow{2}{*}{\Checkmark}} & \multicolumn{1}{c}{\multirow{2}{*}{\Checkmark}} & 90.2\hspace{-1mm} & 74.4\hspace{-1mm} & 60.4\hspace{-1mm} & 17.5\hspace{-1mm} & 66.7\hspace{-1mm} \\
        \multicolumn{1}{c}{} & \multicolumn{1}{c}{} & \multicolumn{1}{c}{}  & 79.5\hspace{-1mm} & 57.7\hspace{-1mm} & 61.8\hspace{-1mm} & 27.4\hspace{-1mm} & 59.6\hspace{-1mm} \\
        \bottomrule
    \end{tabular}%
    }
    \end{minipage}
\end{minipage}

\subsection{Ablation Studies}
To evaluate the effectiveness of our proposed modules and components, we conduct several ablation experiments. We perform ablation experiments on the S2S task to investigate the contribution of the renormalized cosine similarity (RCS) and the Redundant Feature Eliminator (RFE) module. (Table~\ref{tab: ablation}) We perform ablation experiments on the S2R task to investigate the contribution of the RFE module, the spatial Noise Compensator (SNC) module, and the contrastive learning. (Table~\ref{tab: ablation2}) For no contrastive learning (CL) case, training loss $\mathcal{L}^b = \mathcal{L}^b_{align} + \mathcal{L}^b_{cls}$


\noindent\textbf{Effect of RFE and RCS.}
For no RFE case, we use cosine similarity between $\mathbf{f}^g$ and ${\rm \mathbf{F}}^t$ to calculate logits. In table~\ref{tab: ablation}, one can notice that RFE+SNC (line 4) significantly outperforms only SNC (line 3), this phenomenon verifies the conclusion of our analysis that redundant information affects the classification. While only RFE (line 2) can only slightly outperform baseline (line 1) in $\rm NCAcc$. We analyze the reason as follows: The huge domain gap leads to some semantic features available in synthetic data can not be extracted efficiently on real-scanned data. Adding some real-scanned data samples in the principal component extraction stage as prior knowledge may effectively alleviate this symptom.

To further demonstrate the more detailed effectiveness of RFE and RCS, we provide more ablation results in Table~\ref{tab: ablation}. For no Dimensionality Reduction (DR) and no RCS case (line 1), we use cosine similarity between $\mathbf{f}^{2D}$ and $\mathbf{F}^t$ to calculate logits. For with DR and no RCS case (line 2), we use cosine similarity between $\widetilde{\mathbf{f}}^{2D}$ and $\widetilde{\mathbf{F}}^t$ to calculate logits. For DR and renormalized cosine similarity (RCS) case (line 3), we use Eq.\ref{eq5} to calculate logits. 

In Table~\ref{tab: ablation}, it can be observed that if DR is used without the RCS (line 2), The improvement in performance is limited. Conversely, the combination of DR and the RCS (line 3) leads to a significant improvement in performance, particularly in terms of $\rm NCAcc$. We can thus conclude that, without using the RCS, the process of dimensionality reduction leads to inappropriate stretching, which ultimately distorts the semantic information when eliminating redundant information. 

\noindent\textbf{Effect of SNC.}
Let's compare the results of the line 2 with the line 4. On the one hand, there is a slight decrease in $\rm Acc$, which indicates that the 3D module is not as capable as the multi-view model with 2D prior knowledge. It is suitable only for supplementing the information in the 3D FSCIL task. On the other hand, there is a significant increase in the incremental part of the evaluation metric, indicating that the 3D module can complement the global information that the multi-view model lacks and enhance the anti-interference ability.

\begin{minipage}[t]{1.0\textwidth}
\begin{minipage}[h]{0.4\textwidth}
    \makeatletter\def\@captype{table}
    \captionsetup{font=scriptsize}
    \centering
    \caption{Impact of the number of views $N$.}
    \label{view}
    \resizebox{1.0\linewidth}{!}{
    \begin{tabular}{cccccc}
    \toprule[1.2pt]
     $N$ & 0 & 11 & ${\rm NCAcc}\uparrow$ & ${\rm \Delta}\downarrow$ & ${\rm F}\uparrow$ \\ 
     
        \midrule
        \multicolumn{1}{c}{\multirow{2}{*}{4}} & 89.6 & 71.0 & 56.4 & 20.8 & 62.9\\
        \multicolumn{1}{l}{} & 77.6 & 54.5 & 57.5 & 29.8 & 56.0\\
        \midrule
        \multicolumn{1}{c}{\multirow{2}{*}{6}} & 90.0 & 74.6 & 60.0 & 17.1 & 66.9\\
        \multicolumn{1}{l}{} & 79.4 & 57.3 & 59.9 & 27.8 & 58.6\\
        \midrule
        \multicolumn{1}{c}{\multirow{2}{*}{8}} & 90.0 & 74.1 & 60.0 & 17.7 & 66.3\\
        \multicolumn{1}{l}{} & 78.4 & 56.7 & 59.3 & 27.7 & 58.0\\
        \midrule
        \multicolumn{1}{c}{\multirow{2}{*}{10}} & \textbf{90.3} & \textbf{75.3} & \textbf{62.8} & \textbf{16.6} & \textbf{68.5}\\
        \multicolumn{1}{l}{} & \textbf{79.5} & \textbf{57.8} & \textbf{63.3} & \textbf{27.3} & \textbf{60.4}\\
        \midrule
        \multicolumn{1}{c}{\multirow{2}{*}{14}} & 90.0 & 75.0 & 61.8 & 16.7 & 67.8\\
        \multicolumn{1}{l}{} & 79.4 & 57.1 & 60.7 & 28.1 & 58.8\\
    \bottomrule[1.2pt]
    \end{tabular}%
    }
    \end{minipage}
    \begin{minipage}[h]{0.5\textwidth}
   \makeatletter\def\@captype{table}
    \captionsetup{font=scriptsize}
    \centering
    \caption{Impact of the batch size $BS$. \textcolor{red}{Memory refers to the maximum GPU memory consumption during the experiment.}}
    \label{batchsize}
    \resizebox{1.0\linewidth}{!}{
    \begin{tabular}{ccccccc}
    \toprule[1.2pt]
      $BS$ & 0 & 11 & ${\rm NCAcc}\uparrow$ & ${\rm \Delta}\downarrow$ & ${\rm F}\uparrow$ & Memory \\ 
        \midrule
        \multicolumn{1}{c}{\multirow{2}{*}{8}} & 90.0 & 66.1 & 59.6 & 26.5 & 62.7 & \multicolumn{1}{c}{\multirow{2}{*}{5240Mib}}\\
        \multicolumn{1}{l}{} & 78.7 & 56.5 & 59.5 & 28.2 & 58.0 & \multicolumn{1}{c}{}\\
        \midrule
        \multicolumn{1}{c}{\multirow{2}{*}{16}} & 90.0 & 71.8 & 56.7 & 20.2 & 63.4 & \multicolumn{1}{c}{\multirow{2}{*}{7848Mib}}\\
        \multicolumn{1}{l}{} & 79.8 & \textbf{57.4} & 58.2 & 28.1 & 57.8 & \multicolumn{1}{c}{}\\
        \midrule
        \multicolumn{1}{c}{\multirow{2}{*}{32}} & 90.0 & \textbf{74.6} & 60.0 & \textbf{17.1} & 66.9 & \multicolumn{1}{c}{\multirow{2}{*}{12838Mib}}\\
        \multicolumn{1}{l}{} & 79.4 & 57.3 & 59.9 & \textbf{27.8} & 58.6 & \multicolumn{1}{c}{}\\
        \midrule
        \multicolumn{1}{c}{\multirow{2}{*}{64}} & 90.2 & 73.8 & 58.9 & 18.2 & 65.5 & \multicolumn{1}{c}{\multirow{2}{*}{22844Mib}}\\
        \multicolumn{1}{l}{} & 79.7 & 57.2 & 60.0 & 28.2 & 58.6 & \multicolumn{1}{c}{}\\
        \midrule
        \multicolumn{1}{c}{\multirow{2}{*}{128}} & \textbf{90.3} & 72.8 & \textbf{63.7} & 19.6 & \textbf{67.9} & \multicolumn{1}{c}{\multirow{2}{*}{37644Mib}}\\
        \multicolumn{1}{l}{} & \textbf{79.9} & 56.4 & \textbf{64.2} & 29.4 & \textbf{60.0} & \multicolumn{1}{c}{}\\
    \bottomrule[1.2pt]
    \end{tabular}%
    }
    \end{minipage}
\end{minipage}

\begin{minipage}[t]{1.0\textwidth}
    \begin{minipage}[h]{0.6\textwidth}
    \makeatletter\def\@captype{table}
    \captionsetup{font=scriptsize}
    \centering
    \caption{Impact of principal components $M$ of the DR}
    \resizebox{0.9\linewidth}{!}{
    \label{M}
    \begin{tabular}{cccccc}
    \toprule[1.2pt]
      $\rm M$ of the DR & 0 & 11 & ${\rm NCAcc}\uparrow$ & ${\rm \Delta}\downarrow$ & ${\rm F}\uparrow$ \\ 
        \midrule
        \multicolumn{1}{c}{\multirow{2}{*}{127(80\% energy)}} & 89.6 & 72.1 & 52.4 & 19.5 & 60.7\\
        \multicolumn{1}{l}{} & 76.9 & 56.9 & 53.3 & 28.6 & 55.0\\
        \midrule
        \multicolumn{1}{c}{\multirow{2}{*}{188(90\% energy)}} & 90.2 & 72.4 & 55.7 & 19.7 & 63.0\\
        \multicolumn{1}{l}{} & 78.6 & 56.1 & 57.9 & 28.6 & 57.0\\
        \midrule
        \multicolumn{1}{c}{\multirow{2}{*}{242(95\% energy)}} & 90.0 & 74.6 & \textbf{60.0} & \textbf{17.1} & \textbf{66.9}\\
        \multicolumn{1}{l}{} & 79.4 & 57.3 & \textbf{59.9} & 27.8 & \textbf{58.6}\\
        \midrule
        \multicolumn{1}{c}{\multirow{2}{*}{354(99\% energy)}} & 90.5 & 75.0 & 53.7 & \textbf{17.1} & 62.6\\
        \multicolumn{1}{l}{} & 79.9 & \textbf{58.4} & 55.5 & \textbf{26.9} & 56.9\\
        \midrule
        \multicolumn{1}{c}{\multirow{2}{*}{452(99.9\% energy)}} & \textbf{90.6} & \textbf{75.1} & 54.4 & 17.4 & 63.1\\
        \multicolumn{1}{l}{} & \textbf{80.4} & 58.1 & 55.7 & 27.7 & 56.9\\
    \bottomrule[1.2pt]
    \end{tabular}%
    }
    \end{minipage}
    \begin{minipage}[h]{0.35\textwidth}
    \makeatletter\def\@captype{table}
    \captionsetup{font=scriptsize}
    \centering
    \caption{Impact of $\alpha$ and $\beta$}
    \resizebox{1.0\linewidth}{!}{
    \label{alpha}
    \begin{tabular}{cccccccc}
        \toprule[1.2pt]
        $\alpha$ & $\beta$ & 0 & 11 & ${\rm NCAcc}\uparrow$ & ${\rm \Delta}\downarrow$ & ${\rm F}\uparrow$ \\ 
        \midrule
        \multicolumn{1}{c}{\multirow{2}{*}{0.1}} & \multicolumn{1}{c}{\multirow{2}{*}{0.1}} & 89.9 & 70.5 & 56.3 & 21.6 & 62.6\\
        \multicolumn{1}{l}{} & \multicolumn{1}{l}{} & 76.9 & 55.6 & 58.4 & 27.7 & 57.0\\
        \midrule
        \multicolumn{1}{c}{\multirow{2}{*}{0.1}} & \multicolumn{1}{c}{\multirow{2}{*}{0.25}} & 88.8 & 65.2 & 58.1 & 26.6 & 61.4\\
        \multicolumn{1}{l}{} & \multicolumn{1}{l}{} & 75.8 & 52.0 & 60.1 & 31.4 & 55.8\\
        \midrule
        \multicolumn{1}{c}{\multirow{2}{*}{0.25}} & \multicolumn{1}{c}{\multirow{2}{*}{0.1}} & 88.9 & 63.2 & 58.7 & 28.9 & 60.9\\
        \multicolumn{1}{l}{} & \multicolumn{1}{l}{} & 75.0 & 51.1 & 60.3 & 31.9 & 55.3\\
        \midrule
        \multicolumn{1}{c}{\multirow{2}{*}{0.25}} & \multicolumn{1}{c}{\multirow{2}{*}{0.25}} & \textbf{90.2} & \textbf{74.4} & 60.4 & \textbf{17.5} & \textbf{66.7}\\
        \multicolumn{1}{l}{} & \multicolumn{1}{l}{} & \textbf{79.5} & \textbf{57.7} & \textbf{61.8} & 27.4 & \textbf{59.7}\\
        \midrule
        \multicolumn{1}{c}{\multirow{2}{*}{0.25}} & \multicolumn{1}{c}{\multirow{2}{*}{0.5}} & 88.8 & 67.2 & 55.1 & 24.3 & 60.6\\
        \multicolumn{1}{l}{} & \multicolumn{1}{l}{} & 75.4 & 51.4 & 54.5 & 31.1 & 53.0\\
        \midrule
        \multicolumn{1}{c}{\multirow{2}{*}{0.5}} & \multicolumn{1}{c}{\multirow{2}{*}{0.25}} & 88.6 & 64.9 & 52.6 & 26.7 & 58.1\\
        \multicolumn{1}{l}{} & \multicolumn{1}{l}{} & 74.7 & 51.5 & 54.5 & 31.1 & 53.0\\
        \midrule
        \multicolumn{1}{c}{\multirow{2}{*}{0.5}} & \multicolumn{1}{c}{\multirow{2}{*}{0.5}} & 90.0 & 71.8 & \textbf{61.1} & 20.2 & 66.0\\
        \multicolumn{1}{l}{} & \multicolumn{1}{l}{} & 77.6 & 56.4 & 60.1 & \textbf{27.3} & 58.2\\
        \bottomrule[1.2pt]
  \end{tabular}%
    }
    \end{minipage}
\end{minipage}

\noindent\textbf{Hyperparameter sensitivity.}
In Tables \ref{view}, \ref{batchsize}, \ref{M}, and \ref{alpha}, we conduct experiments varying the number of views, batch size, number of principal components in dimensionality reduction (DR), and the values of $\alpha$ and $\beta$ in Eq. \ref{loss3} on the S2R task. \textcolor{red}{An insufficient number of views leads to a lack of semantic information, while an excessive number of views hampers the model's ability to learn class-specific features in few-shot incremental scenarios. As a result, using an intermediate number of views, specifically 10, achieves optimal performance. However, to ensure a fair comparison, we also adopt the default setting of 6 viewpoints as used in PointCLIP~\cite{zhang2022pointclip}.} A larger batch size implies improved learning capabilities, resulting in an increase in $\rm NCAcc$. However, enhanced learning capacity not only facilitates feature dispersion ($\uparrow$) but also leads to a more pronounced compression of the previous feature space ($\downarrow$), causing a nonlinear change in the model's susceptibility to forgetting. \textcolor{red}{The memory capacity of common consumer-grade GPUs, such as the NVIDIA RTX 4090 GPU, is limited to 24GB. Balancing memory cost and performance considerations, we set the batch size to 32 instead of 128.} \textcolor{red}{Too many principal components may fail to eliminate all redundant features, while too few principal components may remove semantic features, both of which disrupt subsequent classification. Therefore, the optimal number of principal components lies in moderation, and we set the number of principal components to 242.} Regarding the selection of $\alpha$ and $\beta$, disparate learning rates between the 2D and 3D branches may cause one branch to become overfitted before the other completes alignment. If $\alpha$ and $\beta$ are too low, they may not effectively accomplish the alignment task, while excessively high values of $\alpha$ and $\beta$ may lead to the compression and excessive smoothing of 2D/3D features, resulting in a decline in model performance. Therefore, we set $\alpha$ and $\beta$ equal to 0.25.

\subsection{Visualization and Analysis}
\begin{figure*}[!t]
  \centering
  \includegraphics[width=0.9\textwidth]{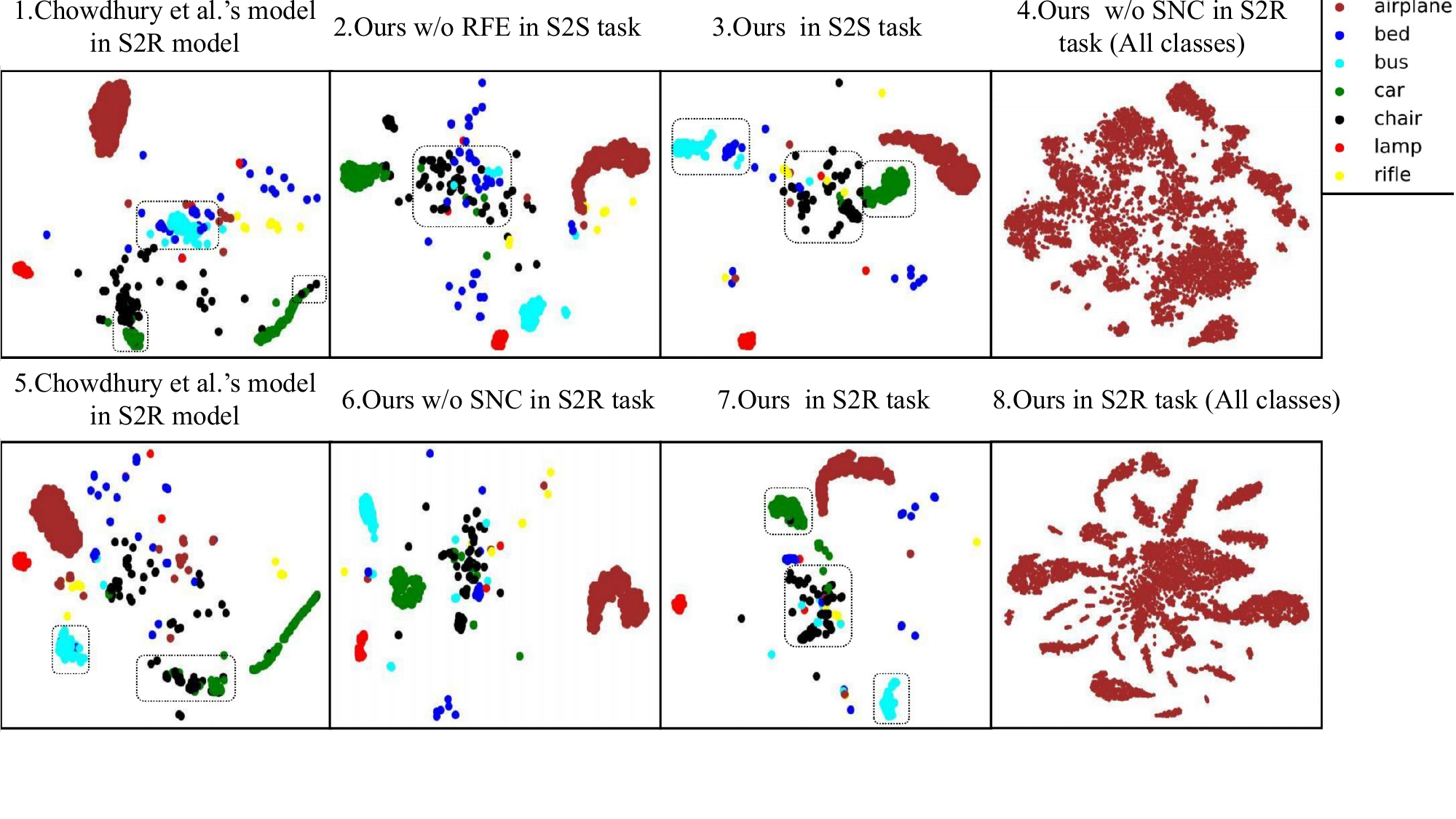}
  \caption{Visualization of experimental results.}
  \label{fig: vis}
\end{figure*}
The visualization results of \textbf{some} classes in the test set are presented in figures 1-3 and 5-7 of Figure~\ref{fig: vis}. Figures 1-3 present the visualizations for \citet{FSCIL-3D}'s model (1), ours w/o RFE (2), and ours (3) in the S2S task. Figures 5-7 similarly display the visualizations for \citet{FSCIL-3D}'s model (5), ours w/o SNC (6), and ours (7) in the S2R task. Figures 4 and 8 present the visualization results of \textbf{all} classes in the S2R test set, with ours w/o SNC on (4) and ours on (8).  These visualizations are generated using t-SNE~\cite{tsne}.

\noindent\textbf{Comparison with \citet{FSCIL-3D}'s}
Comparing our model with \citet{FSCIL-3D}'s, three observations can be concluded: 1) In our model, the intra-class samples appear more compact, indicating that samples within the same class are closer in the feature space. In contrast, the visualizations of \citet{FSCIL-3D}'s model, using chair and bed as examples, exhibit a more scattered distribution. The lack of distinct clustering centers suggests its inability to extract more generalizable features.
2) In our model, there is a notable amount of blank space, indicating that other classes have sufficient feature space for representation. In contrast, the class distributions nearly fill the entire space in the visualizations of \citet{FSCIL-3D}'s model, indicating a propensity for conflicts with unrepresented classes.
3) Considering the displayed classes in Figure~\ref{fig: vis}, our model can accurately distinguish between bed and bus, chair and car, while \citet{FSCIL-3D}'s model demonstrates confusion as indicated by the dashed boxes.

Based on the above observations, we can conclude that the features of our model surpass those extracted by \citet{FSCIL-3D}'s model in aspects of both generalizability and discriminability.

\noindent\textbf{Effect of RFE}
Observing Figure~\ref{fig: vis} (2), we can notice that the bed and chair classes remain relatively dispersed compared to our model. This indicates that the model struggles to extract more generalizable features under the influence of redundant information, resulting in a more scattered distribution.

\noindent\textbf{Effect of SNC}
The distinction between ours w/o SNC (6) and ours (7) is not apparent in specific classes. Nevertheless, by comparing ours w/o SNC (4) with ours (8) in all classes, we can observe that noise greatly affects the discriminability of features (4). The inclusion of supplementary information through SNC effectively alleviates this issue (8).

%% file: sections/5_conclusion.tex
\section{Limitation}
\begin{figure}[t]
  \centering
  \includegraphics[width=1.0\textwidth]{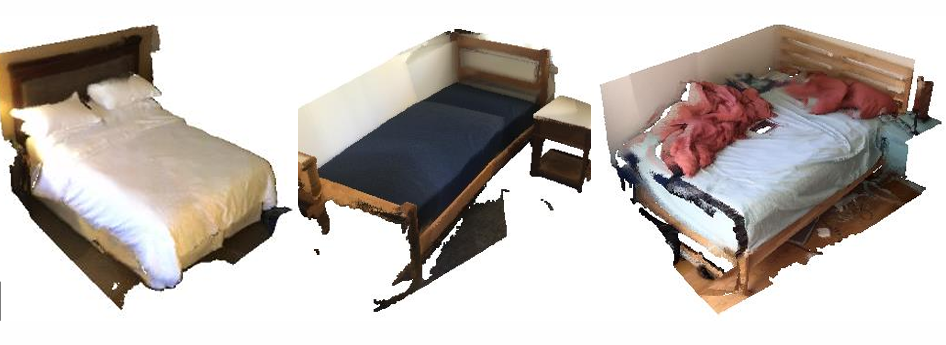}
  \caption{Most views lack informative content. Only the frontal, right-side, and top-down perspectives provide semantic information.}
  \label{fig:failure}
\end{figure}
\textcolor{red}{This work still faces limitations: first, the performance of multi-view fusion is limited when noise is prominent. For example, when point clouds are heavily occluded by walls or floors, as shown in Figure~\ref{fig:failure}, less than half of the views contain semantic information. This lack of 2D information extraction and insufficient multi-view fusion further exacerbates the challenge of few-shot continual learning. Second, insufficient consideration has been given to the complementarity between 2D and 3D features, potentially leading to a suboptimal fusion strategy. For example, the integrated 2D features may mislead the 3D features when the projections of each view cannot extract effective semantic features.}

\section{Conclusion}
\textcolor{red}{In this paper, we proposed FILP-3D, a 3D few-shot class-incremental learning framework with pre-trained V-L models. Specifically, we introduced a V-L pre-trained model CLIP to the 3D FSCIL task, demonstrating its potential to mitigate overfitting and forgetting issues and narrow the domain gap in the 3D FSCIL task. To guarantee that CLIP performs well in the 3D FSCIL task, we proposed a Redundant Feature Eliminator to eliminate redundant features without stretching semantic information and a Spatial Noise Compensator to complement noise information. The two proposed modules are theoretically applicable to other 3D vision tasks utilizing V-L pre-trained models. To comprehensively evaluate models' performance in the 3D FSCIL task, we further proposed an open-source benchmark FSCIL3D-XL, which retains all classes and introduces reasonable evaluation metrics. Extensive experiments demonstrate that our FILP-3D achieves state-of-the-art performance in all available benchmarks. Results from ablation studies and visualization further verify the effectiveness of the proposed components. }

\textcolor{red}{Nonetheless, this work still faces limitations, such as the multi-view fusion lacks robustness in extremely noisy scenarios, and multi-modal feature fusion is suboptimal when 2D features cannot be effectively extracted. In the future, we attempt to leverage 3D features to achieve adaptive view selection, thereby mitigating the impact of uninformative views and extracting richer 2D information.
At the same time, multi-level interactive fusion and cross-modal attention can be used to explore fusion strategies, to facilitate the integration of more comprehensive global features.}